%% file: acl2020.tex
\documentclass[11pt,a4paper]{article}
\usepackage[hyperref]{acl2020}
\usepackage{times}
\usepackage{latexsym}

\usepackage{url}
\usepackage{makecell}
\usepackage{multirow}
\usepackage{graphicx}
\usepackage{caption}
\usepackage{subcaption}
\usepackage{amsmath}
\usepackage{bbm}
\usepackage{booktabs}
\usepackage{algorithmic}
\usepackage[ruled,vlined,linesnumbered]{algorithm2e}
\usepackage{microtype}

\input{math_commands.tex}

\aclfinalcopy 
\def\aclpaperid{1674} 

\newcommand{\hd}{Hard Debias}
\newcommand{\doublehd}{Double-Hard Debias}

\newcommand{\glove}{GloVe}
\newcommand{\hdglove}{Hard-GloVe}
\newcommand{\stronghdglove}{Strong Hard-GloVe}
\newcommand{\gnglove}{GN-GloVe}
\newcommand{\gnaglove}{GN-\glove($w_a$)}
\newcommand{\gpglove}{GP-GloVe}
\newcommand{\gpgnglove}{GP-GN-GloVe}
\newcommand{\doublehdglove}{Double-Hard GloVe}

\newcommand{\wv}{Word2Vec}

\title{Double-Hard Debias:\\ Tailoring Word Embeddings for Gender Bias Mitigation}

\author{
    Tianlu Wang$^1$\thanks{~~This research was conducted during the author's internship at Salesforce Research.} \qquad
    Xi Victoria Lin$^2$ \qquad
    Nazneen Fatema Rajani$^2$\\
{\bf 
    Bryan McCann$^2$ \qquad
    Vicente Ordonez$^1$ \qquad
    Caiming Xiong$^2$
 }
\\
  $^1$University of Virginia \qquad
  \{tw8cb, vicente\}@virginia.edu \\
  $^2$Salesforce Research \qquad
   \{xilin, nazneen.rajani, bmccann, cxiong\}@salesforce.com
}

\begin{document}
\maketitle

\begin{abstract}
Word embeddings derived from human-generated corpora inherit strong gender bias which can be further amplified by downstream models.
Some commonly adopted debiasing approaches, including the seminal \hd~algorithm ~\cite{Bolukbasi2016ManIT}, apply post-processing procedures that project pre-trained word embeddings into a subspace orthogonal to an inferred gender subspace.
We discover that semantic-agnostic corpus regularities such as word frequency captured by the word embeddings negatively impact the performance of these algorithms.
We propose a simple but effective technique, \doublehd, which purifies the word embeddings against such corpus regularities prior to inferring and removing the gender subspace.
Experiments on three bias mitigation benchmarks show that our approach preserves the distributional semantics of the pre-trained word embeddings while reducing gender bias to a significantly larger degree than prior approaches.
\end{abstract}

\section{Introduction}
\input{sections/intro.tex}
\label{sec:intro}

\section{Motivation}
\input{sections/motivation.tex}
\label{sec:motivation}

\section{Method}
\input{sections/double_hard_debias.tex}
\label{sec:discovering}

\section{Experiments}
\label{sec:experiment}
\input{sections/experiment.tex}

\section{Related Work}
\input{sections/related_work.tex}
\label{sec:related_work}

\section{Conclusion}
\input{sections/conclusion.tex}
\label{sec:conclusion}

\bibliography{acl2020}
\bibliographystyle{acl_natbib}

\newpage
\appendix
\section{Appendices}
\input{sections/appendix.tex}

\end{document}

%% file: math_commands.tex
\usepackage{amsmath,amsfonts,bm}

\def\eqref#1{equation~\ref{#1}}

\def\1{\bm{1}}

\def\rvu{{\mathbf{i}}}

\def\rvu{{\mathbf{u}}}

\def\rmC{{\mathbf{C}}}

\def\ow{{\overrightarrow{w}}}

\DeclareMathAlphabet{\mathsfit}{\encodingdefault}{\sfdefault}{m}{sl}
\SetMathAlphabet{\mathsfit}{bold}{\encodingdefault}{\sfdefault}{bx}{n}

\def\gP{{\mathcal{P}}}

\def\gW{{\mathcal{W}}}

\DeclareMathOperator*{\argmin}{arg\,min}

\newcommand{\Realn}[1]{\mathbbm{R}^{#1}}

\usepackage{mathtools}
\newcommand{\defeq}{\vcentcolon=}

%% file: sections/intro.tex
Despite widespread use in natural language processing (NLP) tasks, word embeddings have been criticized for inheriting unintended gender bias from training corpora. \citet{Bolukbasi2016ManIT} highlights that in word2vec embeddings trained on the Google News dataset \citep{Mikolov2013EfficientEO},  ``programmer'' is more closely associated with ``man'' and ``homemaker'' is more closely associated with ``woman''. Such gender bias also propagates to downstream tasks. Studies have shown that coreference resolution systems exhibit gender bias in predictions due to the use of biased word embeddings~\citep{ZWYOC18, rudinger2018gender}. Given the fact that pre-trained word embeddings have been integrated into 
a vast number of NLP models, 
it is important to debias word embeddings to prevent discrimination in NLP systems. 

To mitigate gender bias, prior work have proposed to remove the gender component from pre-trained word embeddings through post-processing~\citep{Bolukbasi2016ManIT}, or to compress the gender information into a few dimensions of the embedding space using a modified training scheme \citep{Zhao2018LearningGW, gp_glove}. We focus on post-hoc gender bias mitigation for two reasons: 1) debiasing via a new training approach is more computationally expensive; and 2) pre-trained biased word embeddings have already been extensively adopted in downstream NLP products and
post-hoc bias mitigation presumably leads to less changes in the model pipeline since it keeps the core components of the original embeddings.

Existing post-processing algorithms, including the seminal \hd~\cite{Bolukbasi2016ManIT}, debias embeddings by removing the component 
that corresponds to a gender direction as defined by a list of gendered words. While \citet{Bolukbasi2016ManIT} demonstrate that such methods alleviate gender bias in word analogy tasks, \citet{Gonen2019LipstickOA} argue that the effectiveness of these efforts is limited, as the gender bias can still be recovered from the geomrtry of the debiased embeddings.

\begin{figure*}[t]
    \begin{subfigure}{0.5\textwidth}
        \centering
        \includegraphics[height=2.7in]{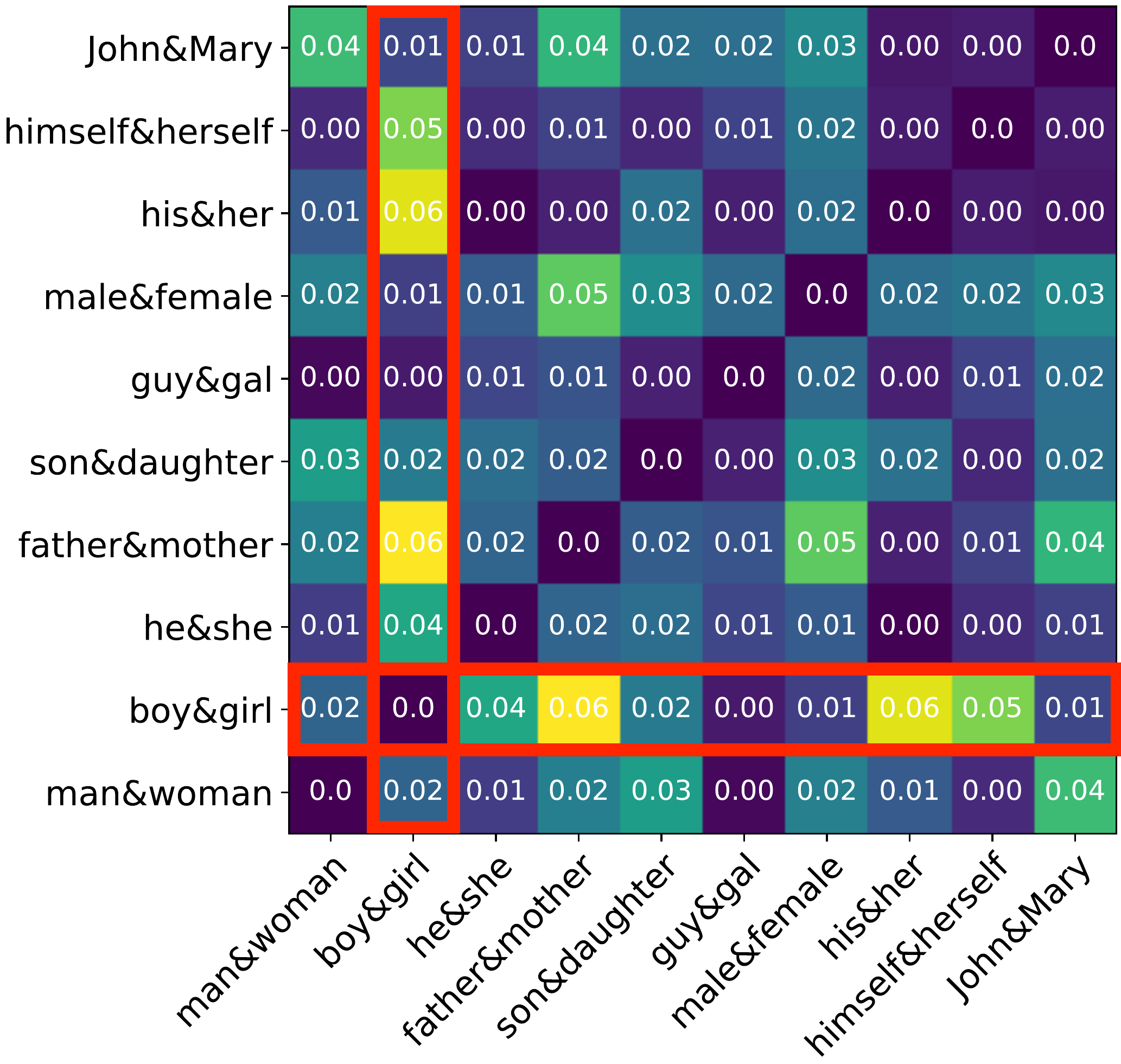}
        \caption{Change the frequency of ``boy''.}
        \label{fig:motivation1}
    \end{subfigure}
    \begin{subfigure}{0.5\textwidth}
        \centering
        \includegraphics[height=2.7in]{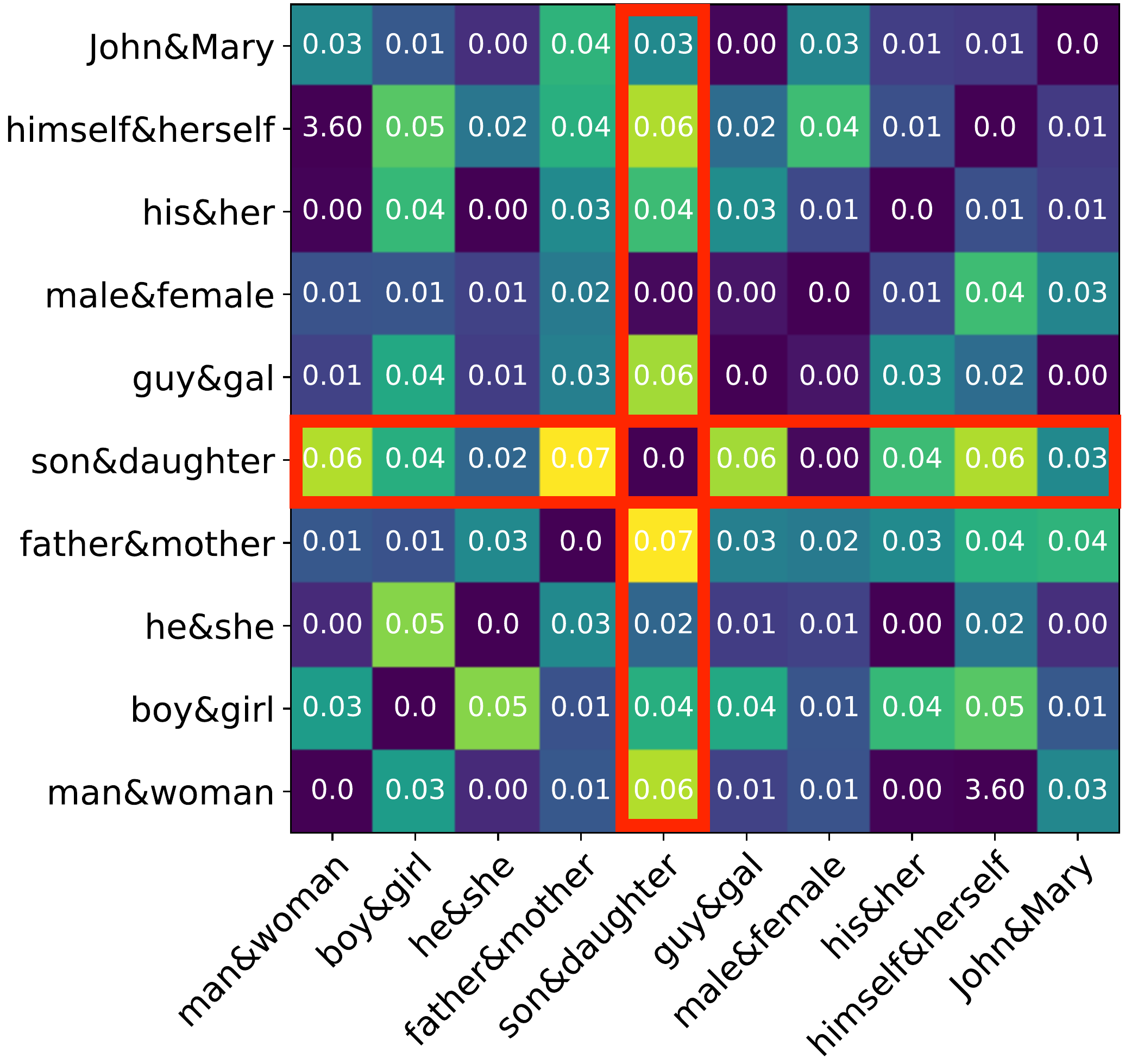}
        \caption{Change the frequency of  ``daughter''.}
        \label{fig:motivation2}
    \end{subfigure}
    \caption{$\Delta$ of cosine similarities between gender difference vectors before / after adjusting the frequency of word $w$. When the frequency of $w$ changes, the cosine similarities between the gender difference vector ($\overrightarrow{v}$) for $w$ and other gender difference vectors exhibits a large change. This demonstrates that frequency statistics for $w$ have a strong influence on the the gender direction represented by $\overrightarrow{v}$.}
    \label{fig:motivation}
\end{figure*}

\textbf{We hypothesize that it is difficult to isolate the gender component of word embeddings} in the manner employed by existing post-processing methods.
For example, \citet{FRAGE, mu2018allbutthetop} show that word frequency significantly impact 
the geometry of word embeddings. Consequently, popular words and rare words cluster 
in different subregions of the embedding space, despite the fact that words in these clusters are not semantically similar.
This can degrade the ability of component-based methods for debiasing gender.

Specifically, recall that \hd~seeks to remove the component of the embeddings corresponding to the gender direction. 
The important assumption made by \hd~is that we can effectively identify and isolate this gender direction.
However, we posit that \textbf{word frequency in the training corpora can twist the gender direction} and limit the effectiveness of \hd.

To this end, we propose a novel debiasing algorithm called \emph{\doublehd}~that builds upon the existing \hd~technique. 
It consists of two steps. 
First, we project word embeddings into an intermediate subspace by subtracting component(s) related to word frequency.
This mitigates the impact of frequency on the gender direction.
Then we apply \hd~to these purified embeddings to mitigate gender bias. 
\citet{mu2018allbutthetop} showed that typically more than one dominant directions in the embedding space encode frequency features. We test the effect of each dominant direction on the debiasing performance and only remove the one(s) that demonstrated the most impact.

We evaluate our proposed debiasing method using a wide range of evaluation techniques. 
According to both representation level evaluation (WEAT test \citep{Caliskan183}, the neighborhood metric \citep{Gonen2019LipstickOA}) and downstream task evaluation (coreference resolution \citep{ZWYOC18}), \doublehd~outperforms all previous debiasing methods. We also evaluate the functionality of debiased embeddings on several benchmark datasets to demonstrate that \doublehd~effectively mitigates gender bias without sacrificing the quality of word embeddings\footnote{Code and data are available at \url{https://github.com/uvavision/Double-Hard-Debias.git}}.

%% file: sections/motivation.tex
Current post-hoc debiasing methods attempt to reduce gender bias in word embeddings by subtracting the component associated with gender from them.
Identifying the gender direction in the word embedding space requires a set of gender word pairs, $\gP$, which consists of ``she \& he'', ``daughter \& son'', etc.
For every pair, for example ``boy \& girl'', the difference vector of the two embeddings is expected to approximately capture the gender direction:
\begin{equation}
    \overrightarrow{v}_{boy, girl} = \overrightarrow{w}_{boy}-\overrightarrow{w}_{girl}
\label{eq:difference vector}
\end{equation}
\citet{Bolukbasi2016ManIT} computes the first principal component of ten such difference vectors and use that to define the gender direction.\footnote{The complete definition of $\gP$ is: ``woman \& man'', ``girl \& boy'', ``she \& he'', ``mother \& father'', ``daughter \& son'', ``gal \& guy'', ``female \& male'', ``her \& his'', ``herself \& himself'', and ``Mary \& John''~\citep{Bolukbasi2016ManIT}.}

Recent works \citep{mu2018allbutthetop, FRAGE} show that word frequency in a training corpus can degrade the quality of word embeddings. By carefully removing such frequency features, existing word embeddings can achieve higher performance on several benchmarks after fine-tuning.
We hypothesize that such word frequency statistics also interferes with the components of the word embeddings associated with gender.
In other words, frequency-based features learned by word embedding algorithms act as harmful noise in the previously proposed debiasing techniques.

To verify this, we first retrain \glove~\cite{pennington-etal-2014-glove} embeddings on the one billion English word benchmark~\citep{chelba2013one} following previous work~\cite{Zhao2018LearningGW,gp_glove}.
We obtain ten difference vectors for the gendered pairs in $\gP$ and compute pairwise cosine similarity. 
This gives a similarity matrix $\mathcal{S}$ in which $\mathcal{S}_{p_i, p_j}$ denotes the cosine similarity between difference vectors $\overrightarrow{v}_{pair_i}$ and $\overrightarrow{v}_{pair_j}$.

We then select a specific word pair, e.g. ``boy'' \& ``girl'', and augment the corpus by sampling sentences containing the word ``boy'' twice. In this way, we produce a new training corpus with altered word frequency statistics for ``boy''. 
The context around the token remains the same so that changes to the other components are negligible. 
We retrain \glove~with this augmented corpus and get a set of new offset vectors for the gendered pairs $\gP$.
We also compute a second similarity matrix $\mathcal{S}'$ where $\mathcal{S}_{p_i, p_j}'$ denotes the cosine similarity between difference vectors $\overrightarrow{v}_{pair_i}'$ and $\overrightarrow{v}_{pair_j}'$.

By comparing these two similarity matrices, we analyze the effect of changing word frequency statistics on gender direction.
Note that the offset vectors are designed for approximating the gender direction, thus we focus on the changes in offset vectors.
Because statistics were altered for ``boy'', we focus on the difference vector $\overrightarrow{v}_{boy, girl}$ and make two observations.
First, the norm of $\overrightarrow{v}_{boy, girl}$ has a $5.8\%$ relative change while the norms of other difference vectors show much smaller changes.
For example, the norm of $\overrightarrow{v}_{man, woman}$ only changes by $1.8\%$.
Second, the cosine similarities between $\overrightarrow{v}_{boy, girl}$ and other difference vectors also show more significant change, as highlighted by the red bounding box in Figure \ref{fig:motivation1}. 
As we can see, the frequency change of ``boy'' leads to deviation of the gender direction captured by $\overrightarrow{v}_{boy, girl}$. 
We observe similar phenomenon when we change the frequency of the word ``daughter'' and present these results in Figure \ref{fig:motivation2}. 

Based on these observations, we conclude that word frequency plays an important role in gender debiasing despite being overlooked by previous works.

%% file: sections/double_hard_debias.tex
In this section, we first summarize the terminology that will be used throughout the rest of the paper, briefly review the \hd~method, and provide background on the neighborhood evaluation metric. Then we introduce our proposed method: \doublehd. 

\subsection{Preliminary Definitions}
\label{sec:preliminary}
Let $W$ be the vocabulary of the word embeddings we aim to debias.
The set of word embeddings contains a vector $\ow\in\Realn{n}$ for each word $w\in W$.
A subspace $B$ is defined by $k$ orthogonal unit vectors $B=\{b_1,\dots,b_k\}\in\Realn{d}$. 
We denote the projection of vector $v$ on $B$ by 
\begin{equation}
v_{B}=\sum_{j=1}^k(v\cdot b_j)b_j.
\end{equation}

Following~\cite{Bolukbasi2016ManIT}, we assume there is a set of gender neutral words $N\subset W$, such as ``doctor'' and ``teacher'', which by definition are not specific to any gender.
We also assume a pre-defined set of $n$ male-female word pairs $D_1, D_2, \dots, D_n\subset W$, where the main difference between each pair of words captures \textit{gender}. 

{\bf Hard Debias.} 
The \hd~algorithm first identifies a subspace that captures gender bias. Let 
\begin{equation}
\mu_{i}\defeq\sum_{w\in D_i} \overrightarrow{w}/|D_i|.    
\end{equation}
The bias subspace $B$ is the first $k$ ($\geq 1$) rows of SVD($\rmC$), where   
    \begin{equation}
        \rmC\defeq\sum\limits_{i=1}^{m}\sum\limits_{w\in D_i}(\overrightarrow{w} - \mu_{i})^{T}(\overrightarrow{w} - \mu_{i})/|D_{i}|
    \end{equation}
Following the original implementation of \citet{Bolukbasi2016ManIT}, we set $k=1$. As a result the subspace $B$ is simply a gender direction.\footnote{\citet{Bolukbasi2016ManIT} normalize all embeddings. However, we found it is unnecessary in our experiments. This is also mentioned in \citet{ethayarajh2019understanding}}

\hd~then neutralizes the word embeddings by transforming each $\ow$ such that every word $w\in N$ has zero projection in the gender subspace. For each word $w\in N$, we re-embed $\ow$:
    \begin{equation}
        \ow\defeq \ow - \ow_B
    \end{equation}

{\bf Neighborhood Metric.}
The Neighborhood Metric proposed by \cite{Gonen2019LipstickOA} is a bias measurement that does not rely on any specific gender direction.
To do so it looks into similarities between words. The bias of a word is the proportion of words with the same gender bias polarity among its nearest neighboring words. 

We selected $k$ of the most biased male and females words according to the cosine similarity of their embedding and the gender direction computed using the word embeddings prior to bias mitigation.
We use $W_m$ and $W_f$ to denote the male and female biased words, respectively.
For $w_i\in W_m$, we assign a ground truth gender label $g_i = 0$. 
For $w_i\in W_f$, $g_i = 1$. 
Then we run KMeans ($k = 2$) to cluster the embeddings of selected words $\hat{g_i} = KMeans(\ow_i)$, and compute the alignment score $a$ with respect to the assigned ground truth gender labels:
\begin{equation}
\label{eq:alignemnt-score}
    a = \frac{1}{2k}\sum_{i=1}^{2k}\mathbbm{1}{[\hat{g_i}==g_i]}
\end{equation}
We set $a = \max(a, 1-a)$.
Thus, a value of $0.5$ in this metric indicates perfectly unbiased word embeddings (i.e. the words are randomly clustered), and a value closer to $1$ indicates stronger gender bias. 

\subsection{Double-Hard Debiasing}

\begin{algorithm}[t]
\SetAlgoLined
\SetKwInOut{Input}{Input}\SetKwInOut{Output}{Output}
\Input{Word embeddings: $\{\ow\in \Realn{d}, w\in\gW\}$\\
Male biased words set: $W_m$\\
Female biased words set: $W_f$}
$S_{debias} = []$

Decentralize $\ow$: $\mu \leftarrow \frac{1}{|\mathcal{V}|}\sum_{w\in \mathcal{V}}\ow$, for each $\ow\in\gW$, $\Tilde{w} \leftarrow \ow - \mu$;

Compute principal components by PCA: $\{\rvu_1\hdots\rvu_d\}$ $\leftarrow$ PCA(\{$\Tilde{w}$, $w\in\gW$\})\;

//discover the frequency directions

\For{$i = 1$ to d}{
  $w_m^{\prime} \leftarrow \Tilde{w_m} - (\rvu_i^Tw_m)\rvu_i$\;
  $w_f^{\prime} \leftarrow \Tilde{w_f} - (\rvu_i^Tw_f)\rvu_i$\;
  $\hat{w}_m \leftarrow HardDebias(w_m^{\prime})$\;
  $\hat{w}_f \leftarrow HardDebias(w_f^{\prime})$\;
  $output = KMeans([\hat{w}_m \hat{w}_f])$\;
  $a$ = eval(output, $W_m$, $W_f$)\;
  $S_{debias}$.append($a$)\;
 }
 
$k = \argmin_i S_{debias}$\;

// remove component on frequency direction

$w^{\prime} \leftarrow \Tilde{w} - (\rvu_k^Tw)\rvu_k$\;

// remove components on gender direction

$\hat{w} \leftarrow HardDebias(w^{\prime})$\;

\Output{Debiased word embeddings:\\
\{$\hat{w}\in \mathbb{R}^{d}, w\in\gW$\}
}
 \caption{Double-Hard Debias.}
 \label{alg:double-hard-debias}
\end{algorithm}

\begin{figure}[t]
        \centering
        \includegraphics[height=1.2in]{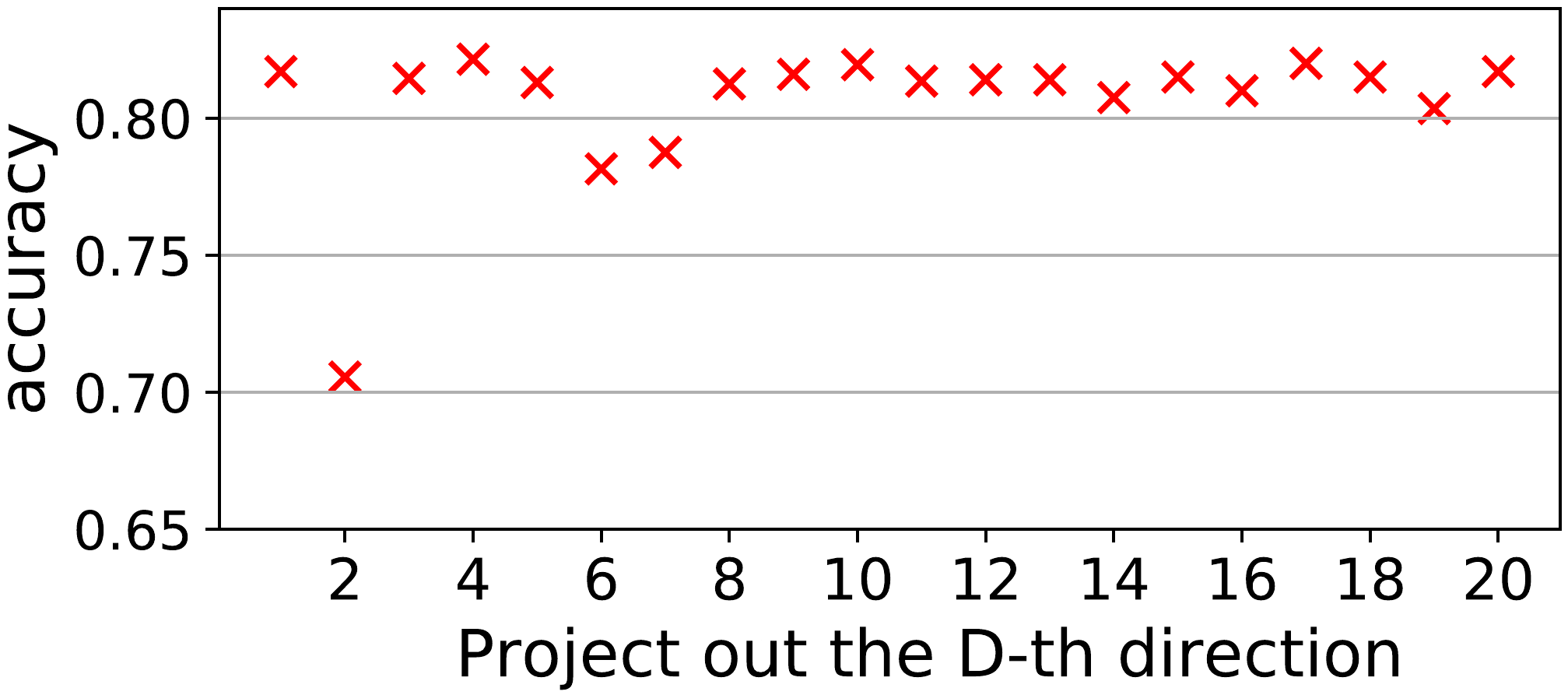}
    \caption{Clustering accuracy after projecting out D-th dominating direction and applying Hard Debias. Lower accuracy indicates less bias.}
    \label{fig:discovering}
\end{figure}
According to~\citet{mu2018allbutthetop}, the most statistically dominant directions of word embeddings encode word frequency to a significant extent.
~\citet{mu2018allbutthetop} removes these frequency features by centralizing and subtracting components along the top $D$ dominant directions from the original word embeddings.
These post-processed embedddings achieve better performance on several benchmark tasks, including word similarity, concept categorization, and word analogy. It is also suggested that setting $D$ near $d/100$ provides maximum benefit, where $d$ is the dimension of a word embedding.

We speculate that most the dominant directions also affect the geometry of the gender space.
To address this, we use the aforementioned clustering experiment to identify whether a direction contains frequency features that alter the gender direction.
  
More specifically, we first pick the top biased words ($500$ male and $500$ female)
identified using the original \glove~embeddings. We then apply PCA to all their word embeddings and take the top principal components as candidate directions to drop. 
For every candidate direction $\mathbf{u}$, we project the embeddings into a space that is orthogonal to $\mathbf{u}$.
In this intermediate subspace, we apply \hd~and get debiased embeddings. 
Next, we cluster the debiased embeddings of these words and compute the gender alignment accuracy (Eq.~\ref{eq:alignemnt-score}).
This indicates whether projecting away direction $\mathbf{u}$ improves the debiasing performance. 
Algorithm \ref{alg:double-hard-debias} shows the details of our method in full.
  
We found that for \glove~embeddings pre-trained on Wikipedia dataset, elimination of the projection along the second principal component significantly decreases the clustering accuracy.
This translates to better debiasing results, as shown in Figure \ref{fig:discovering}. 
We further demonstrate the effectiveness of our method for debaising using other evaluation metrics in Section \ref{sec:experiment}.

%% file: sections/experiment.tex
In this section, we compare our proposed method with other debiasing algorithms and test the functionality of these debiased embeddings on word analogy and concept categorization task. 
Experimental results demonstrate that our method effectively reduces bias to a larger extent without degrading the quality of word embeddings.

\subsection{Dataset}
\label{sec: dataset}
We use 300-dimensional \glove~ \cite{pennington-etal-2014-glove} \footnote{Experiments on Word2Vec are included in the appendix.} embeddings pre-trained on the 2017 January dump of English Wikipedia\footnote{\url{https://github.com/uclanlp/gn\_glove}}, containing $322,636$ unique words. To identify the gender direction, we use $10$ pairs of definitional gender words compiled by  \cite{Bolukbasi2016ManIT}\footnote{\url{https://github.com/tolga-b/debiaswe}}.

\subsection{Baselines}

We compare our proposed method against the following baselines:

\vspace{2mm}
\noindent \textbf{\glove:} the pre-trained \glove~embeddings on Wikipedia dataset described in \ref{sec: dataset}. \glove~is widely used in various NLP applications. This is a non-debiased baseline for comparision.

\vspace{2mm}
\noindent \textbf{\gnglove:} We use debiased Gender-Neutral \gnglove~embeddings released by the original authors \cite{Zhao2018LearningGW}. \gnglove~restricts gender information in certain dimensions while neutralizing the rest dimensions. 

\vspace{2mm}
\noindent \textbf{\gnaglove:} We exclude the gender dimensions from \gnglove. This baseline tries to completely remove gender.

\vspace{2mm}
\noindent \textbf{\gpglove:} We use debiased embeddings released by the original authors \cite{gp_glove}. Gender-preserving Debiasing attempts to preserve non-discriminative gender information, while removing stereotypical gender bias.

\vspace{2mm}
\noindent \textbf{\gpgnglove:}: This baseline applies Gender-preserving Debiasing on already debaised GN-\glove~embeddings. We also use debiased embeddings provided by authors.

\vspace{2mm}
\noindent \textbf{\hdglove:} We apply \hd~introduced in \cite{Bolukbasi2016ManIT} on \glove~embeddings. Following the implementation provided by original authors, we debias netural words and preserve the gender specific words.

\vspace{2mm}
\noindent \textbf{\stronghdglove:} A variant of \hd~where we debias all words instead of avoiding gender specific words. This seeks to entirely remove gender from \glove~embeddings.

\vspace{2mm}
\noindent \textbf{\doublehdglove:} We debias the pre-trained \glove~embeddings by our proposed \doublehd~method.

\begin{table*}[t]
\centering
\small
\begin{tabular}{c|c|cccc|cccc}
\toprule
\bf Embeddings & \makecell{\bf OntoNotes} &  \bf PRO-1 & \bf ANTI-1 & \bf Avg-1 & \bf $|$Diff-1 $|$ & \bf PRO-2 & \bf ANTI-2 & \bf Avg-2 & \bf $|$Diff-2 $|$\\
\midrule
\glove & $\bf 66.5$ & $77.7$& $48.2$& $\bf 62.9$& $29.0$ & $82.7$& $67.5$& $75.1$& $15.2$\\
\midrule 
\gnglove & $66.1$ & $68.4$& $56.5$& $62.5$& $12.0$ & $78.2$& $71.3$& $74.7$& $6.9$\\

\gnaglove & $66.4$ & $66.7$& $56.6$& $61.6$& $10.2$ & $79.0$& $72.3$& $75.7$& $6.7$\\

\midrule

\gpglove & $66.1$ & $72.0$& $52.0$& $62.0$& $20.0$ & $78.5$& $70.0$& $74.3$& $8.6$\\

\gpgnglove & $66.3$ & $70.0$& $54.5$& $62.0$& $15.0$ & $79.9$& $70.7$& $75.3$& $9.2$\\

\midrule

\hdglove & $66.2$ & $72.3$& $52.7$& $62.6$& $19.7$ & $80.6$& $78.3$& $79.4$& $2.3$\\

\stronghdglove & $66.0$ & $69.0$& $58.6$& $63.8$& $10.4$ & $82.2$& $78.6$& $80.4$& $3.6$\\

\midrule

\doublehdglove & $66.4$ & $66.0$& $58.3$& $62.2$& $\bf 7.7$ & $85.4$& $84.5$& $\bf 85.0$& $\bf 0.9$\\
\bottomrule
\end{tabular}
\caption{F1 score (\%) of coreference systems on OntoNotes test set and WinoBias dataset. $|$Diff $|$ represents the performance gap between pro-stereotype (PRO) subset and anti-stereotype (ANTI) subset. Coreference system trained on our Double-Hard \glove~embeddings has the smallest $|$Diff $|$ values, suggesting less gender bias.}
\label{tab:coref}
\end{table*}

\subsection{Evaluation of Debiasing Performance}
We demonstrate the effectiveness of our debiasing method for downstream applications and according to general embedding level evaluations. 

\subsubsection{Debiasing in Downstream Applications}
\textbf{Coreference Resolution.} 
Coreference resolution aims at identifying noun phrases referring to the same entity. \citet{ZWYOC18} identified gender bias in modern coreference systems, e.g. ``doctor'' is prone to be linked to ``he''. They also introduce a new benchmark dataset WinoBias, to study gender bias in coreference systems.

WinoBias provides sentences following two prototypical templates. Each type of sentences can be divided into a pro-stereotype (PRO) subset and a antistereotype (ANTI) subset. In the PRO subset, gender pronouns refer to professions dominated by the same gender. For example, in sentence ``The physician hired the secretary because he was overwhelmed with clients.'',  ``he'' refers to ``physician'', which is consistent with societal stereotype. On the other hand, the ANTI subset consists of same sentences, but the opposite gender pronouns. As such, ``he'' is replaced by ``she'' in the aforementioned example. The hypothesis is that gender cues may distract a coreference model. We consider a system to be gender biased if it performs better in  pro-stereotypical scenarios than in anti-stereotypical scenarios. 

We train an end-to-end coreference resolution model~\citep{DBLP:conf/emnlp/LeeHLZ17} with different word embeddings on OntoNotes 5.0 training set and report the performance on WinoBias dataset. 
Results are presented in Table\ref{tab:coref}. Note that absolute performance difference (Diff) between the PRO set and ANTI set connects with gender bias. A smaller Diff value indicates a less biased coreference system. We can see that on both types of sentences in WinoBias, \doublehdglove~achieves the smallest Diff compared to other baselines. This demonstrates the efficacy of our method.
Meanwhile, \doublehdglove~maintains comparable performance as \glove~on OntoNotes test set, showing that our method preserves the utility of word embeddings.
It is also worth noting that by reducing gender bias, \doublehdglove~can significantly improve the average performance on type-2 sentences, from $75.1\%$ (\glove) to $85.0\%$. 

\subsubsection{Debiasing at Embedding Level}
\textbf{The Word Embeddings Association Test (WEAT).} WEAT is a permutation test used to measure the bias in word embeddins. We consider male names and females names as attribute sets and compute the differential association of two sets of target words\footnote{All word lists are from \citet{Caliskan183}. Because \glove embeddings are uncased, we use lower cased people names and replace ``bill'' with ``tom'' to avoid ambiguity.} and the gender attribute sets. We report effect sizes ($d$) and p-values ($p$) in Table\ref{tab:WEAT}. The effect size is a normalized measure of how separated the two distributions are.  A higher value of effect size indicates larger bias between target words with regard to gender. p-values denote if the bias is significant. A high p-value (larger than $0.05$) indicates the bias is insignificant. We refer readers to \citet{Caliskan183} for more details.

As shown in Table \ref{tab:WEAT}, across different target words sets, \doublehdglove~consistently outperforms other debiased embeddings. For Career \& Family and Science \& Arts, \doublehdglove~reaches the lowest effect size, for the latter one, \doublehdglove~successfully makes the bias insignificant (p-value $> 0.05$). Note that in WEAT test, some debiasing methods run the risk of amplifying gender bias, e.g. for Math \& Arts words, the bias is significant in GN-\glove~while it is insignificant in original \glove~embeddings. Such concern does not occur in \doublehdglove.

\begin{table*}[t]
\centering
\small
\begin{tabular}{ccccccc}
\toprule
\multirow{2}{*}{\textbf{Embeddings}} & \multicolumn{2}{c}{\bf Career \& Family } & \multicolumn{2}{c}{\bf Math \& Arts} & \multicolumn{2}{c}{\bf Science \& Arts}\\
 & \bf $d$ & \bf $p$ &  \bf $d$ & \bf $p$ & \bf $d$ & \bf $p$\\
\midrule
\glove & $1.81$ & $0.0$& $0.55$ & $0.14$ & $0.88$& $0.04$\\
\midrule
GN-\glove & $1.82$ & $0.0$& $1.21$ & $6\mathrm{e}^{-3}$ & $1.02$& $0.02$\\
\gnaglove & $1.76$ & $0.0$& $1.43$ & $1\mathrm{e}^{-3}$ & $1.02$& $0.02$\\
\midrule
GP-\glove & $1.81$ & $0.0$& $0.87$ & $0.04$ & $0.91$& $0.03$\\
GP-GN-\glove & $1.80$ & $0.0$& $1.42$ & $1\mathrm{e}^{-3}$ & $1.04$& $0.01$\\
\midrule
Hard-\glove & $1.55$ & $2\mathrm{e}^{-4}$& $0.07$ & $0.44$ & $ 0.16$& $0.62$\\
Strong Hard-\glove & $1.55$ & $2\mathrm{e}^{-4}$& $0.07$ & $0.44$ & $ 0.16$& $0.62$\\
\midrule
Double-Hard \glove & $1.53$ & $2\mathrm{e}^{-4}$& $0.09$ & $0.57$ & $ 0.15$& $0.61$\\
\bottomrule
\end{tabular}
\caption{WEAT test of embeddings before/after Debiasing. The bias is insignificant when p-value, $p > 0.05$. Lower effective size ($d$) indicates less gender bias. Significant gender bias related to Career \& Family and Science \& Arts words is effectively reduced by \doublehdglove. Note for Math \& Arts words, gender bias is insignificant in original \glove.}
\label{tab:WEAT}
\end{table*}

\vspace{2mm}
\textbf{Neighborhood Metric.} \cite{Gonen2019LipstickOA} introduces a neighborhood metric based on clustering. As described in Sec \ref{sec:preliminary}, We take the top $k$ most biased words according to their cosine similarity with gender direction in the original \glove~embedding space\footnote{To be fair, we exclude all gender specific words used in debiasing, so \hdglove~and \stronghdglove~have same acurracy performance in Table \ref{tab:bias_eval}}. We then run k-Means to cluster them into two clusters and compute the alignment accuracy with respect to gender, results are presented in Table~\ref{tab:bias_eval}. We recall that in this metric, a accuracy value closer to $0.5$ indicates less biased word embeddings.

\begin{figure*}
\begin{subfigure}{.25\textwidth}
  \centering
  \includegraphics[width=\linewidth]{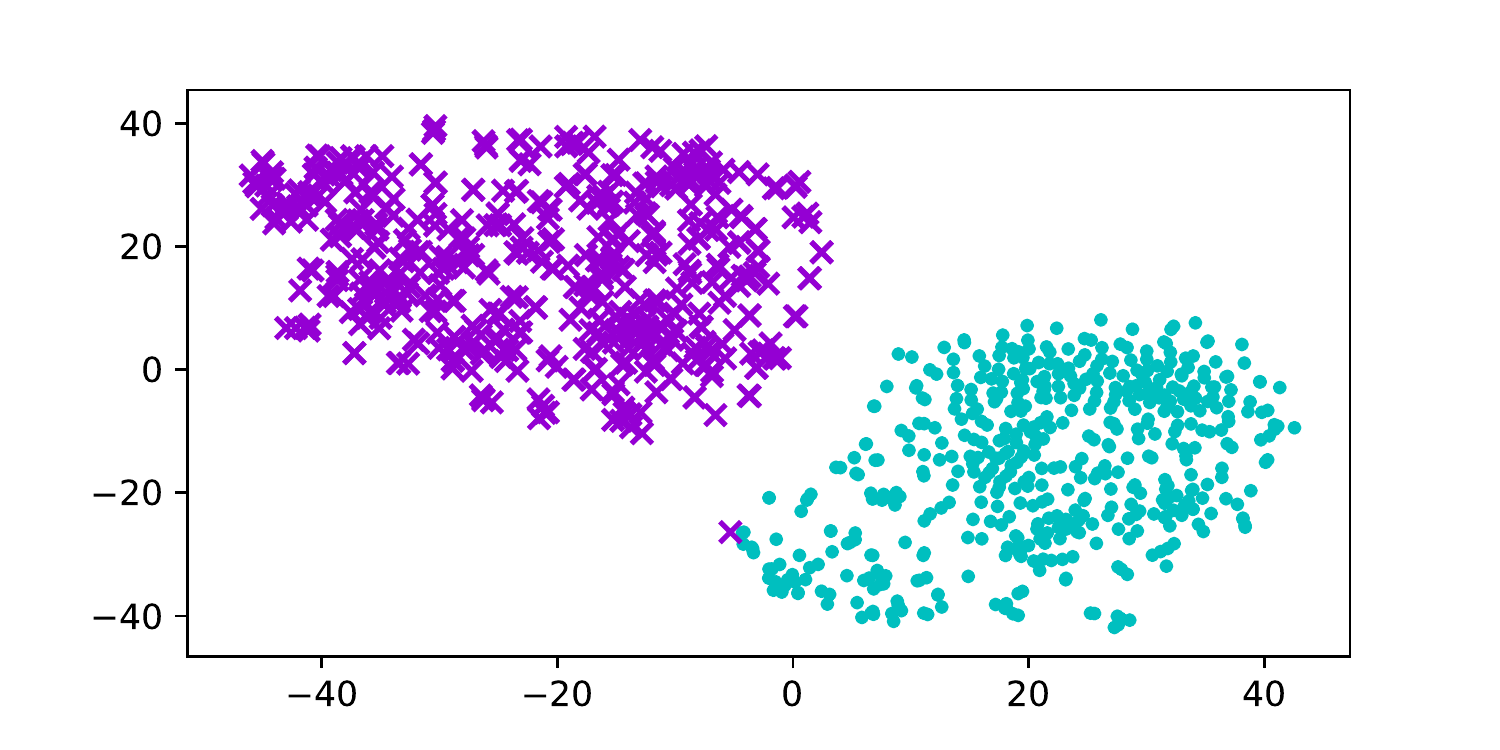}
  \caption{\glove}
  \label{fig:glove}
\end{subfigure}%
\begin{subfigure}{.25\textwidth}
  \centering
  \includegraphics[width=\linewidth]{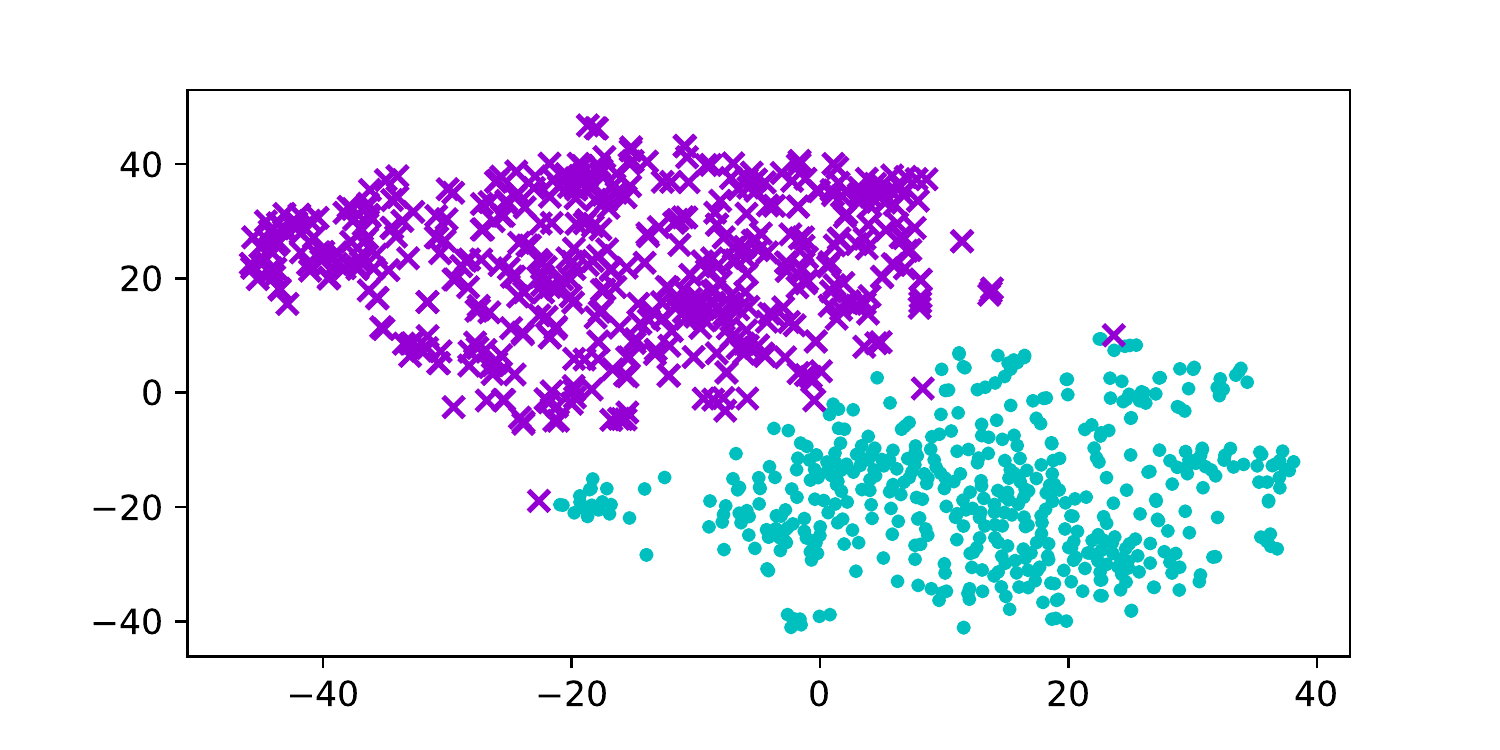}
  \caption{GN-\glove}
  \label{fig:gn_glove}
\end{subfigure}
\begin{subfigure}{.25\textwidth}
  \centering
  \includegraphics[width=\linewidth]{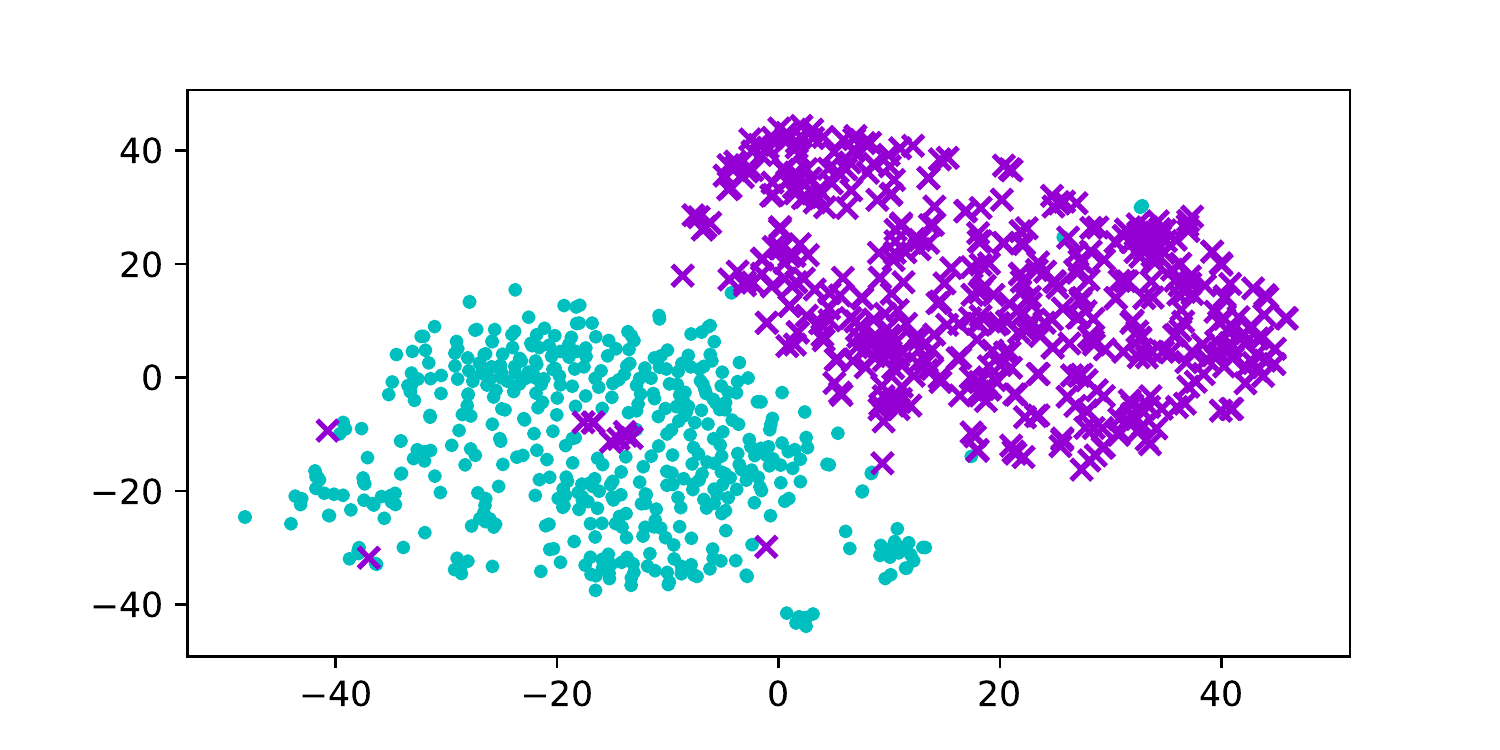}
  \caption{\gnaglove}
  \label{fig:gn_a_glove}
\end{subfigure}%
\begin{subfigure}{.25\textwidth}
  \centering
  \includegraphics[width=\linewidth]{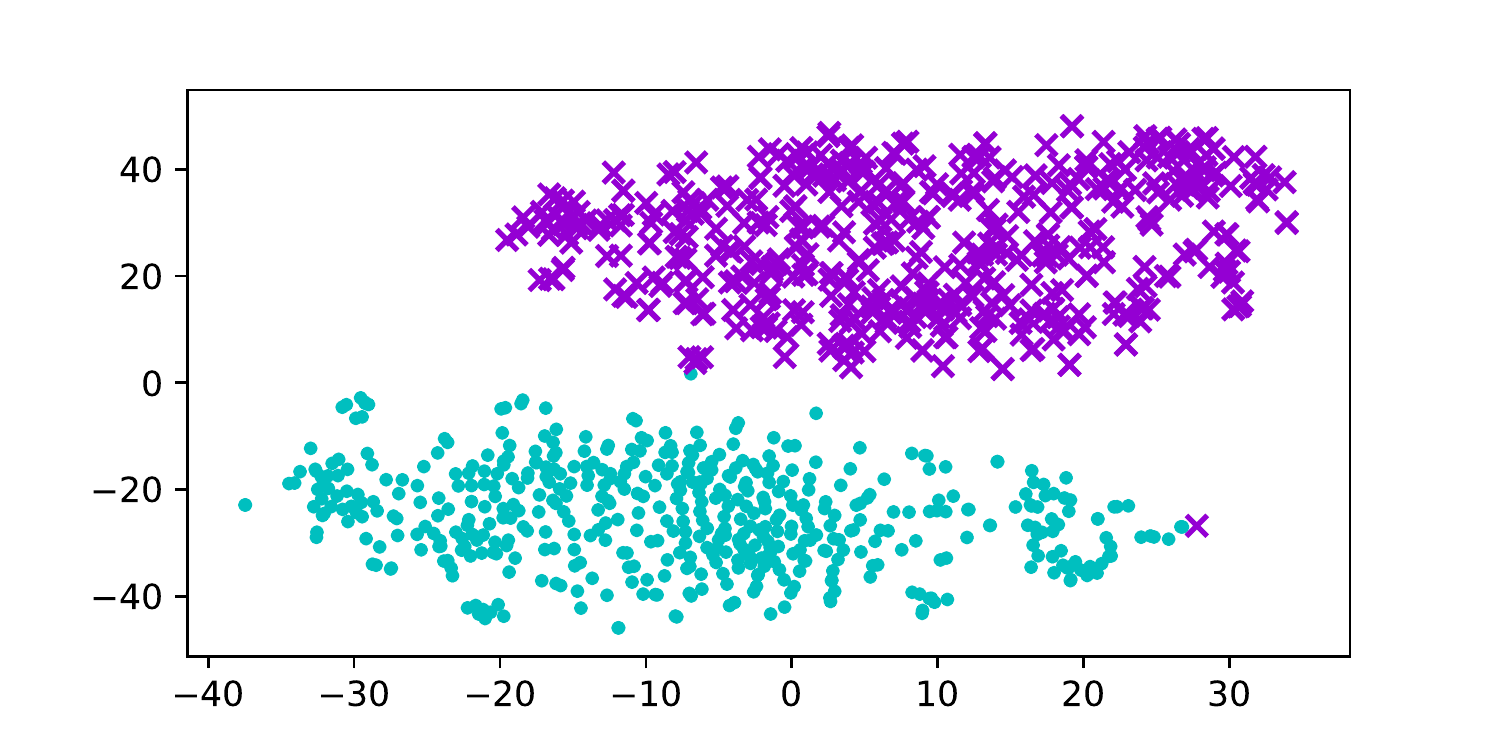}
  \caption{GP-\glove}
  \label{fig:gp_glove}
\end{subfigure}
\medskip
\begin{subfigure}{.25\textwidth}
  \centering
  \includegraphics[width=\linewidth]{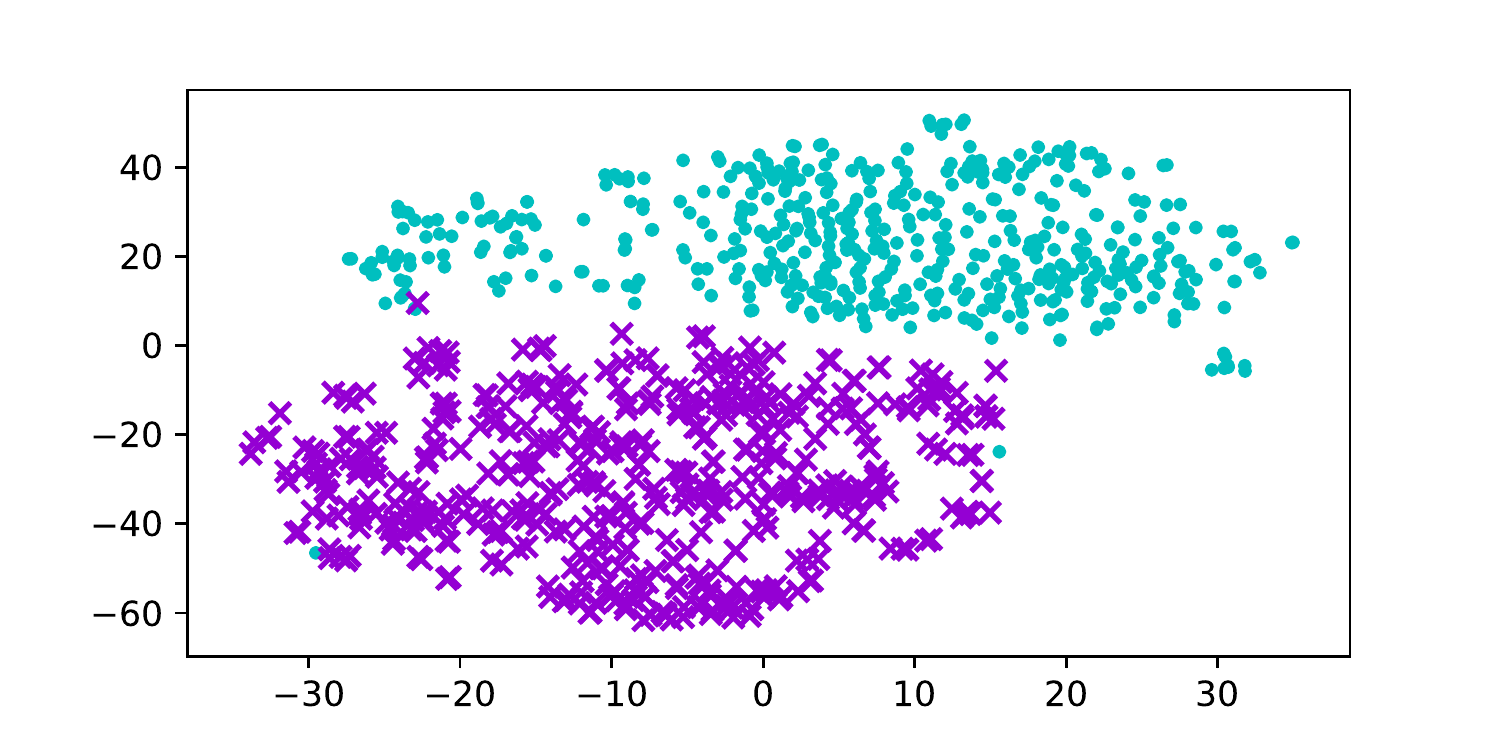}
  \caption{GP-GN-\glove}
  \label{fig:gp_gn_glove}
\end{subfigure}%
\begin{subfigure}{.25\textwidth}
  \centering
  \includegraphics[width=\linewidth]{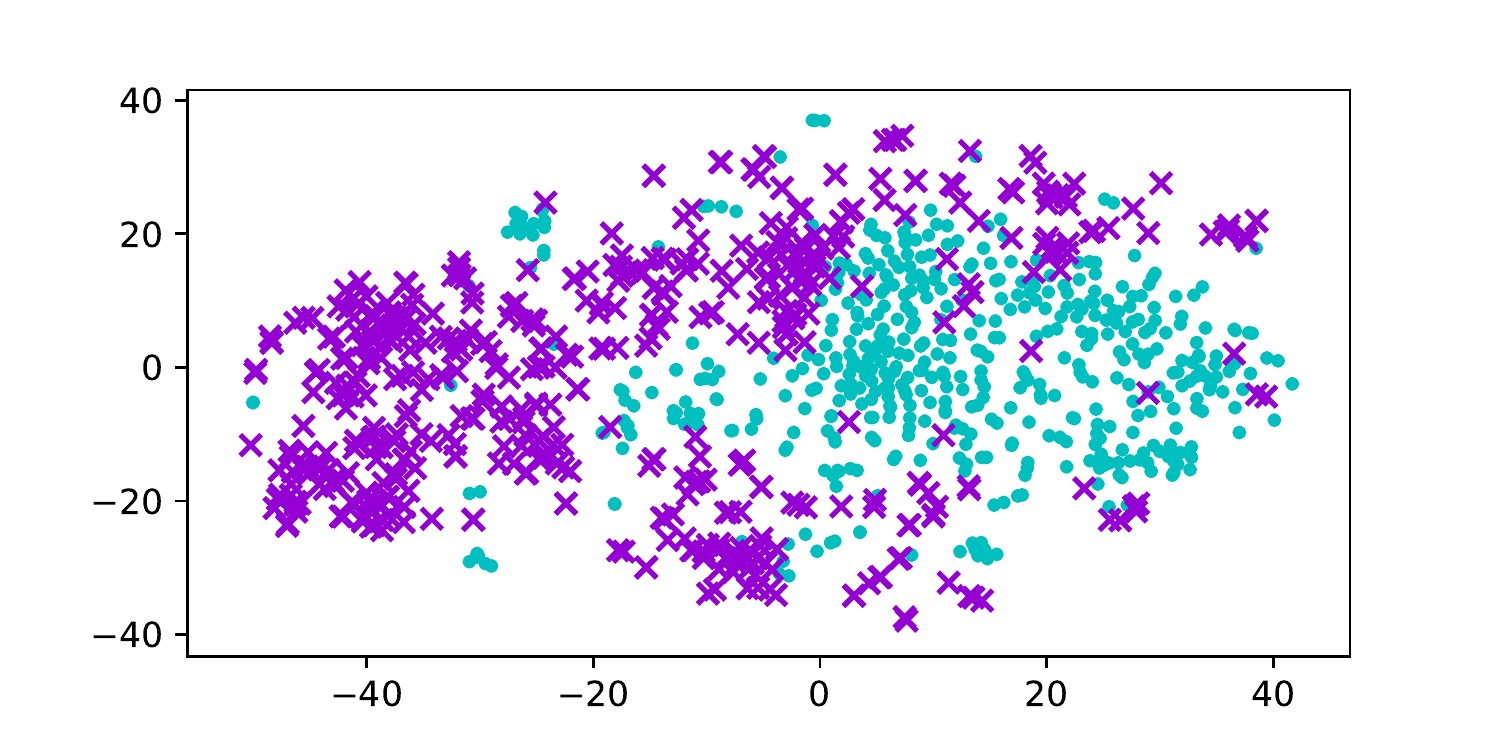}
  \caption{Hard-\glove}
  \label{fig:hd_glove}
\end{subfigure}
\begin{subfigure}{.25\textwidth}
  \centering
  \includegraphics[width=\linewidth]{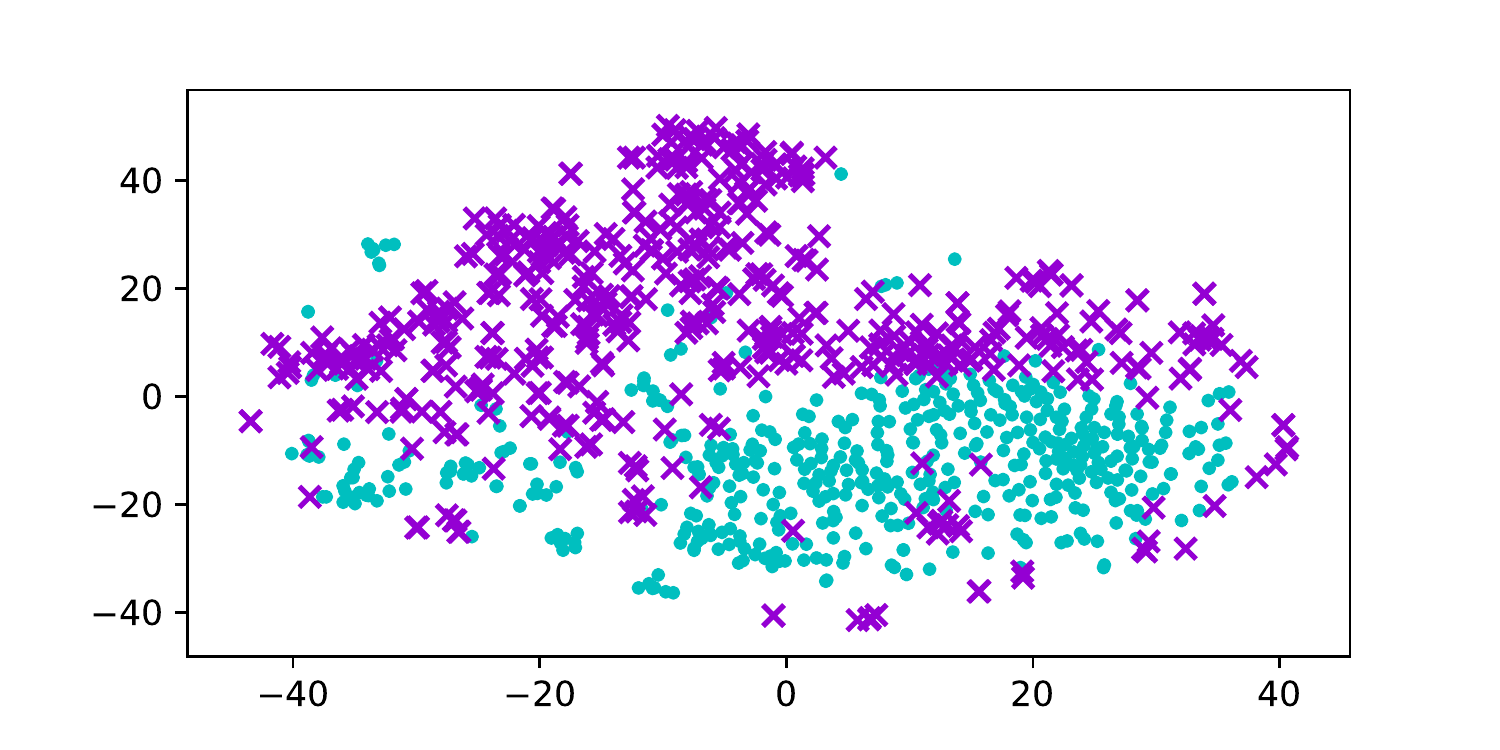}
  \caption{Strong Hard-\glove}
  \label{fig:strong_hd_glove}
\end{subfigure}%
\begin{subfigure}{.25\textwidth}
  \centering
  \includegraphics[width=\linewidth]{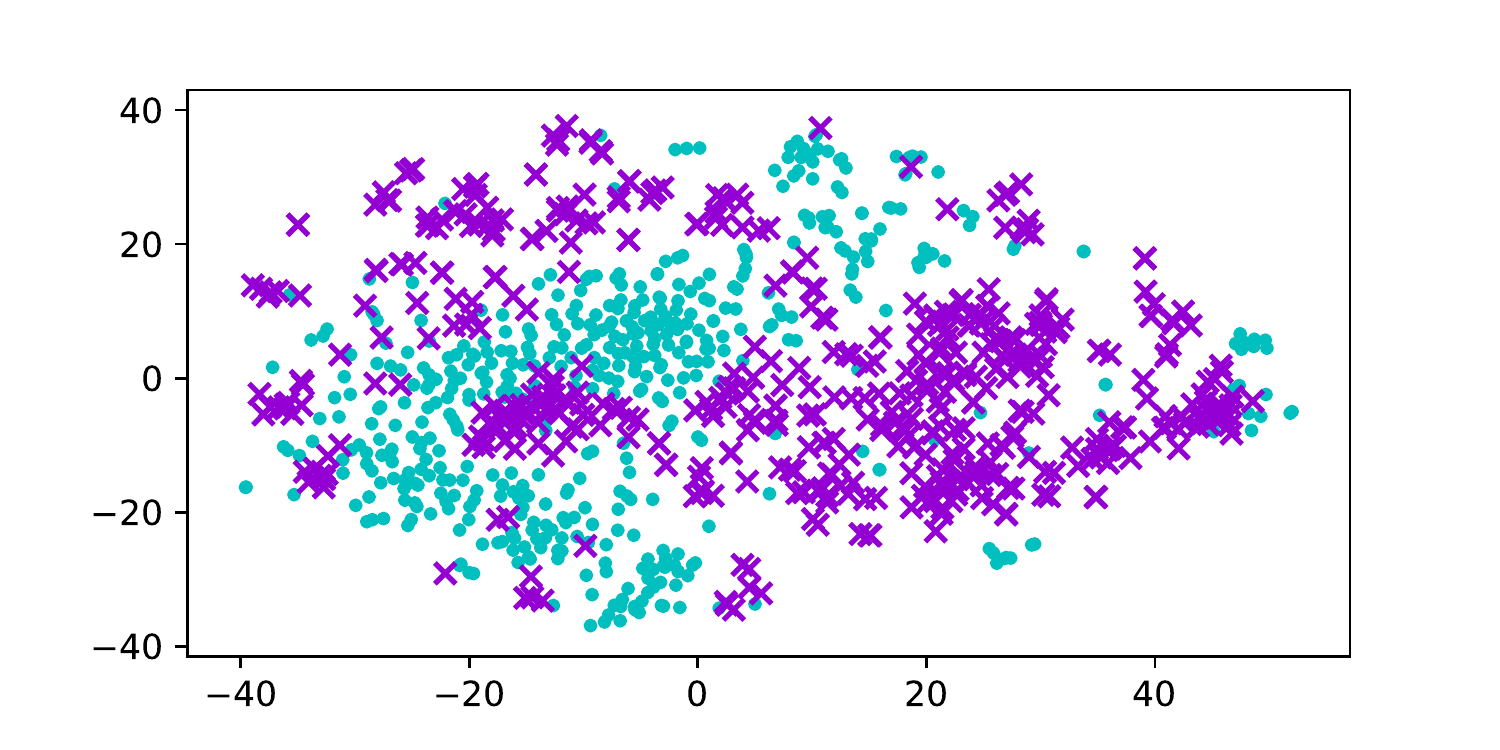}
  \caption{\doublehdglove}
  \label{fig:double_hd_glove}
\end{subfigure}
\caption{tSNE visualization of top $500$ most male and female embeddings. \doublehdglove~mixes up two groups to the maximum extent, showing less gender information is encoded.}
\label{fig:tsne}
\end{figure*}

Using the original \glove~embeddings, k-Means can accurately cluster selected words into a male group and a female group, suggesting the presence of a strong bias. \hd~is able to reduce bias in some degree while other baselines appear to be less effective. \doublehdglove~achieves the lowest accuracy across experiments clustering top 100/500/1000 biased words, demonstrating that the proposed technique effectively reduce gender bias. We also conduct tSNE \citep{vanDerMaaten2008} projection for all baseline embeddings. As shown in Figure~\ref{fig:tsne}, original non-debiased \glove~ embeddings are clearly projected to different regions. \doublehdglove~mixes up male and female embeddings to the maximum extent compared to other baselines, showing less gender information can be captured after debiasing.

\begin{table}[h]
\centering
\small
\begin{tabular}{ccccc}
\toprule
\textbf{Embeddings} & \textbf{Top 100} & \textbf{Top 500} & \textbf{Top 1000}\\
\midrule
\glove & $100.0$ & $100.0$ & $100.0$\\
\midrule
\gnglove & $100.0$ & $100.0$ & $99.9$\\
\gnaglove & $100.0$ & $99.7$ & $88.5$\\
\midrule
\gpglove & $100.0$ & $100.0$ & $100.0$\\
\gpgnglove & $100.0$ & $100.0$ & $99.4$\\
\midrule
(Strong) Hard \glove & $59.0$ & $62.1$ & $68.1$\\
\midrule
Double-Hard \glove & $\bf 51.5$ & $\bf 55.5$& $\bf 59.5$\\
\bottomrule
\end{tabular}
\caption{Clustering Accuracy (\%) of top 100/500/1000 male and female words. Lower accuracy means less gender cues can be captured. \doublehdglove~ consistently achieves the lowest accuracy.}
\label{tab:bias_eval}
\end{table}

\subsection{Analysis of Retaining Word Semantics}

\begin{table*}[t]
\centering
\small
\begin{tabular}{c|cccc|cccc}
\toprule
\multirow{2}{*}{\bf Embeddings}& \multicolumn{4}{c|}{\bf Analogy} &\multicolumn{4}{c}{\bf Concept Categorization}
\\ 
 & Sem & Syn & Total & MSR & AP & ESSLI & Battig & BLESS\\
\midrule
\glove & $80.5$ & $62.8$ & $70.8$ & $ 54.2$& 
$55.6$& $ 72.7$& $51.2$& $81.0$\\
\midrule
GN-\glove & $77.7$ & $61.6$ & $68.9$ & $51.9$& 
$56.9$& $70.5$& $ 49.5$& $ 85.0$\\
\gnaglove & $77.7$ & $61.6$ & $68.9$ & $51.9$& 
$56.9$& $75.0$& $ 51.3$& $ 82.5$\\
\midrule
GP-\glove & $80.6$ & $61.7$ & $70.3$ & $51.3$& 
$56.1$& $75.0$& $ 49.0$& $ 78.5$\\
GP-GN-\glove & $77.7$ & $61.7$ & $68.9$ & $51.8$& 
$61.1$& $72.7$& $ 50.9$& $ 77.5$\\
\midrule
Hard-\glove  & $80.3$ & $62.5$ & $70.6$ & $54.0$& 
$62.3$& $79.5$& $50.0$& $84.5$\\
Strong Hard-\glove  & $78.6$ & $62.4$ & $69.8$ & $53.9$& 
$64.1$& $79.5$& $49.2$& $84.5$\\
\midrule
Double-Hard \glove & $ 80.9$ & $61.6$ & $70.4$ & $53.8$& 
$ 59.6$& $72.7$& $46.7$& $79.5$\\
\midrule
\end{tabular}
\caption{Results of word embeddings on word analogy and concept categorization benchmark datasets. Performance (x100) is measured in accuracy and purity, respectively. On both tasks, there is no significant degradation of performance due to applying the proposed method.}
\label{tab:analogy_catrgorization}
\end{table*}

\textbf{Word Analogy.} Given three words $A$, $B$ and $C$, the analogy task is to find word $D$ such that ``$A$ is to $B$ as $C$ is to $D$''. In our experiments, $D$ is the word that maximize the cosine similarity between $D$ and $C-A+B$.
We evaluate all non-debiased and debiased embeddings on the MSR \citep{mikolov-etal-2013-linguistic} word analogy task, which contains $8000$ syntactic questions, and on a second Google word analogy \citep{Mikolov2013EfficientEO} dataset that contains $19,544$ (\textbf{Total}) questions, including $8,869$ semantic (\textbf{Sem}) and $10,675$ syntactic (\textbf{Syn}) questions. 
The evaluation metric is the percentage of questions for which the correct answer is assigned the maximum score by the algorithm. 
Results are shown in Table\ref{tab:analogy_catrgorization}. \doublehdglove~achieves comparable good results as \glove~ and slightly outperforms some other debiased embeddings. This proves that \doublehd~is capable of preserving proximity among words. 

\textbf{Concept Categorization.} The goal of concept categorization is to cluster a set of words into different categorical subsets. For example, ``sandwich'' and ``hotdog'' are both food and ``dog'' and ``cat'' are animals. The clustering performance is evaluated in terms of purity \citep{manning2008introduction} - the fraction of the total number of the words that are correctly classified. Experiments are conducted on four benchmark datasets: the Almuhareb-Poesio (AP) dataset \citep{DBLP:phd/ethos/Almuhareb06}; the ESSLLI 2008 \citep{essli}; the Battig 1969 set \citep{Battig1969-BATCNO} and the BLESS dataset \citep{Baroni:2011:WBD:2140490.2140491}. We run classical Kmeans algorithm with fixed $k$. Across four datasets, the performance of \doublehdglove~is on a par with \glove~embeddings, showing that the proposed debiasing method preserves useful semantic information in word embeddings. Full results can be found in Table\ref{tab:analogy_catrgorization}.

%% file: sections/related_work.tex
{\bf Gender Bias in Word Embeddings.} Word embeddings have been criticized for carrying gender bias. \citet{Bolukbasi2016ManIT} show that  word2vec \citep{word2vec} embeddings trained on the Google News dataset exhibit occupational stereotypes, e.g. ``programmer'' is closer to ``man'' and ``homemaker'' is closer to ``woman''. More recent works \citep{genderbiaselmo, BERTbias, DBLP:journals/corr/abs-1904-08783} demonstrate that contextualized word embeddings also inherit gender bias.  

Gender bias in word embeddings also propagate to downstream tasks, which substantially affects predictions. \citet{ZWYOC18} show that coreference systems tend to link occupations to their stereotypical gender, e.g. linking ``doctor'' to ``he'' and ``nurse'' to ``she''. \citet{stanovsky2019evaluating} observe that popular industrial and academic machine translation systems are prone to gender biased translation errors.

Recently, ~\citet{vig2020causal} proposed causal mediation analysis as a way to interpret and analyze gender bias in neural models.

\vspace{2mm}
\noindent {\bf Debiasing Word Embeddings.} For contextualized embeddings, existing works propose task-specific debiasing methods, while in this paper we focus on more generic ones. To mitigate gender bias,~\citet{ZWYOC18} propose a new training approach which explicitly restricts gender information in certain dimensions during training. While this method separates gender information from embeddings, retraining word embeddings on massive corpus requires an undesirably large amount of resources.~\citet{gp_glove} tackles this problem by adopting an encoder-decoder model to re-embed word embeddings. This can be applied to existing pre-trained embeddings, but it still requires train different encoder-decoders for different embeddings.

\citet{Bolukbasi2016ManIT} introduce a more simple and direct post-processing method which zeros out the component along the gender direction. 
This method reduces gender bias to some degree, however,~\citet{Gonen2019LipstickOA} present a series of experiments to show that they are far from delivering gender-neutral embeddings. Our work builds on top of \citet{Bolukbasi2016ManIT}. We discover the important factor -- word frequency -- that limits the effectiveness of existing methods. By carefully eliminating the effect of word frequency, our method is able to significantly improve debiasing performance.

%% file: sections/conclusion.tex
We have discovered that simple changes in word frequency statistics can have an undesirable impact on the debiasing methods used to remove gender bias from word embeddings. 
Though word frequency statistics have until now been neglected in previous gender bias reduction work, we propose Double-Hard Debias, which mitigates the negative effects that word frequency features can have on debiasing algorithms.
We experiment on several benchmarks and demonstrate that our Double-Hard Debias is more effective on gender bias reduction than other methods while also preserving the quality of word embeddings suitable for the downstream applications and embedding-based word analogy tasks. 
While we have shown that this method significantly reduces gender bias while preserving quality, we hope that this work encourages further research into debiasing along other dimensions of word embeddings in the future.

%% file: sections/appendix.tex
\begin{figure}[h]
        \centering
        \includegraphics[height=1.2in]{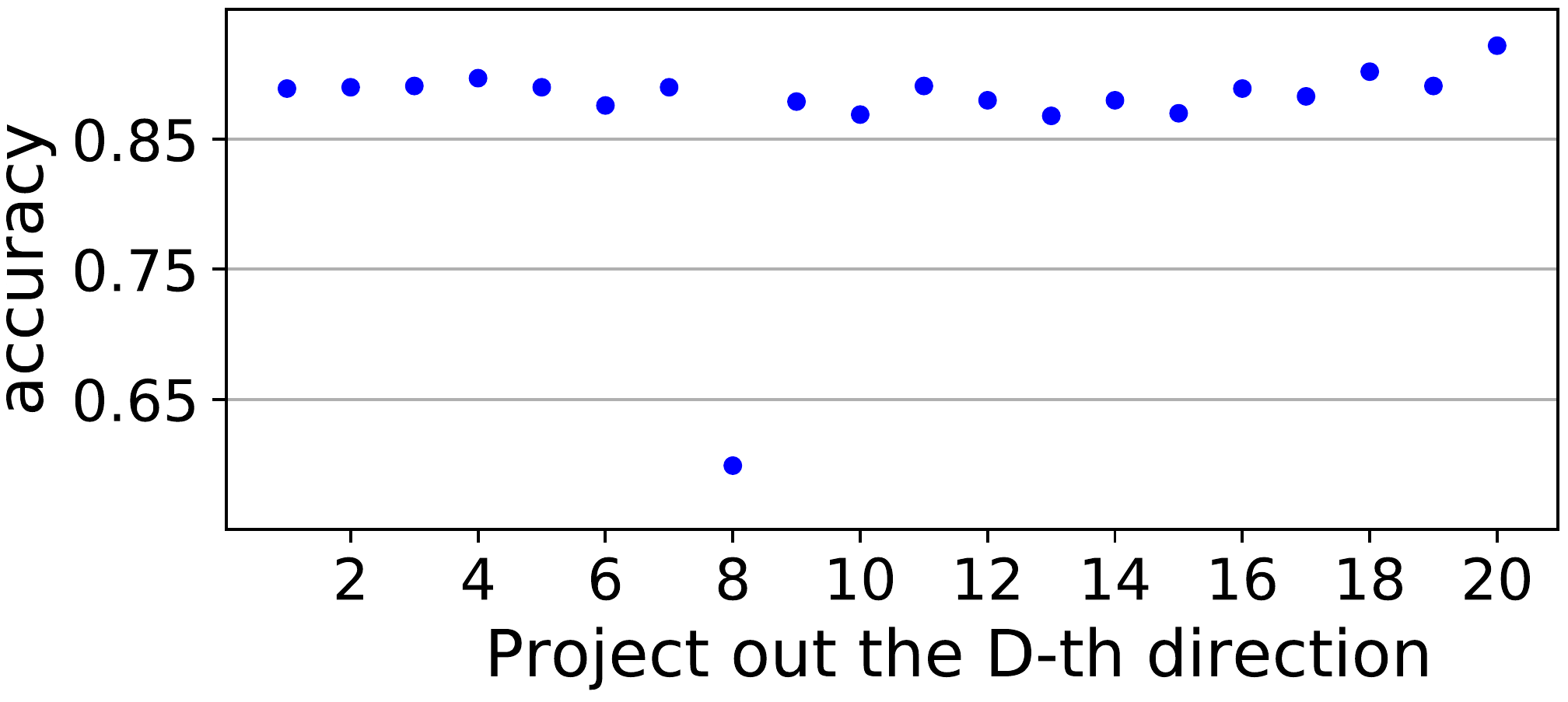}
    \caption{Clustering accuracy after projecting out D-th dominating direction and applying Hard Debias. Lower accuracy indicates less bias.}
    \label{fig:discovering-w2v}
\end{figure}

\begin{table}[h]
\centering
\small
\begin{tabular}{ccccc}
\toprule
\textbf{Embeddings} & \textbf{Top 100} & \textbf{Top 500} & \textbf{Top 1000}\\
\midrule
\wv & $100.0$ & $99.3$ & $99.3$\\
Hard-\wv & $79.5$ & $74.3$ & $79.8$\\
\midrule
\scriptsize{Double-Hard \wv} & $\bf 71.0$ & $\bf 52.3$& $\bf 56.7$\\
\bottomrule
\end{tabular}
\caption{Clustering Accuracy(\%) of top 100/500/1000 male and female words. Lower accuracy means less gender cues captured. Double-Hard \wv~ consistently achieves the lowest accuracy.}
\label{tab:bias_eval_wv}
\end{table}

\begin{figure}[h]
\centering
\begin{subfigure}{.46\textwidth}
  \centering
  \includegraphics[width=\linewidth]{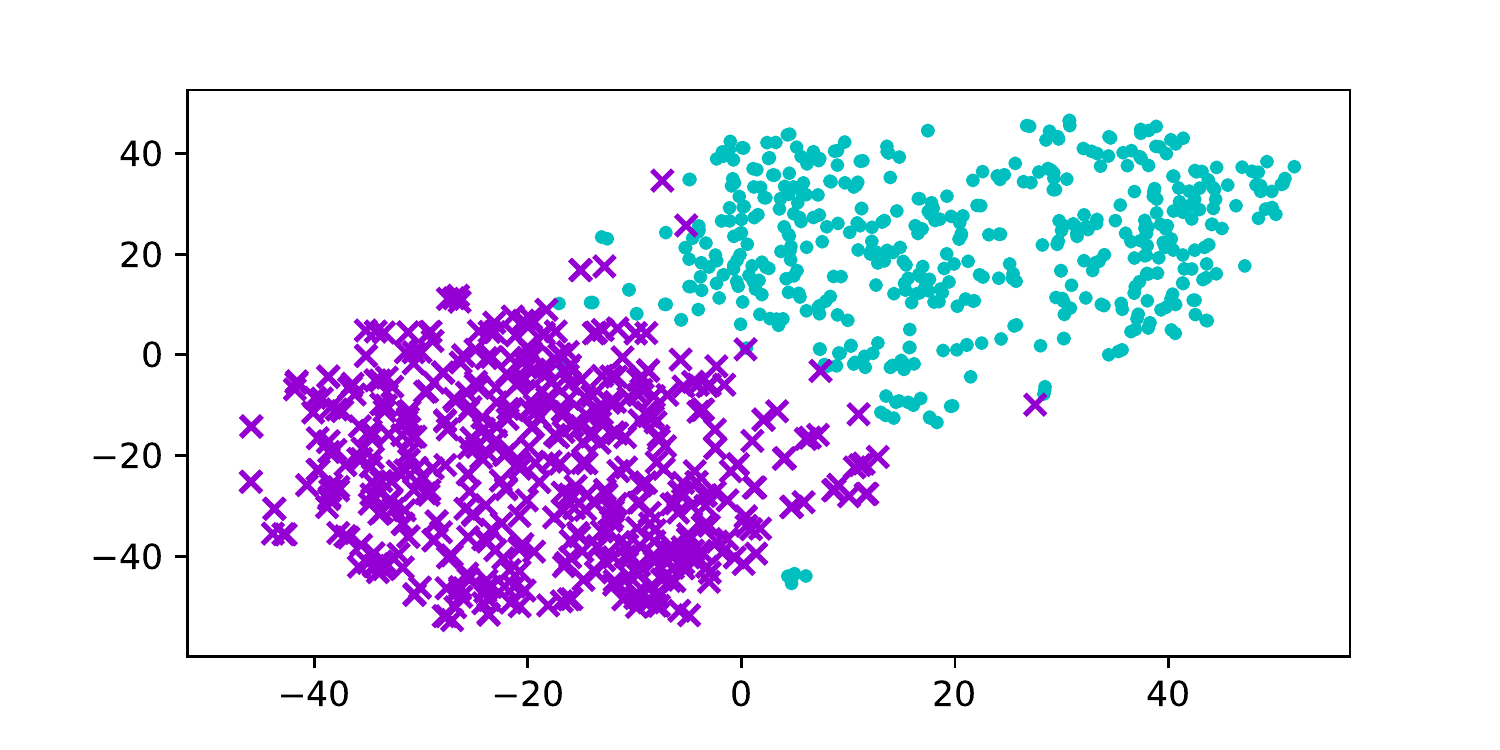}
  \caption{\wv}
  \label{fig:wv}
\end{subfigure}%
\newline
\begin{subfigure}{.46\textwidth}
  \centering
  \includegraphics[width=\linewidth]{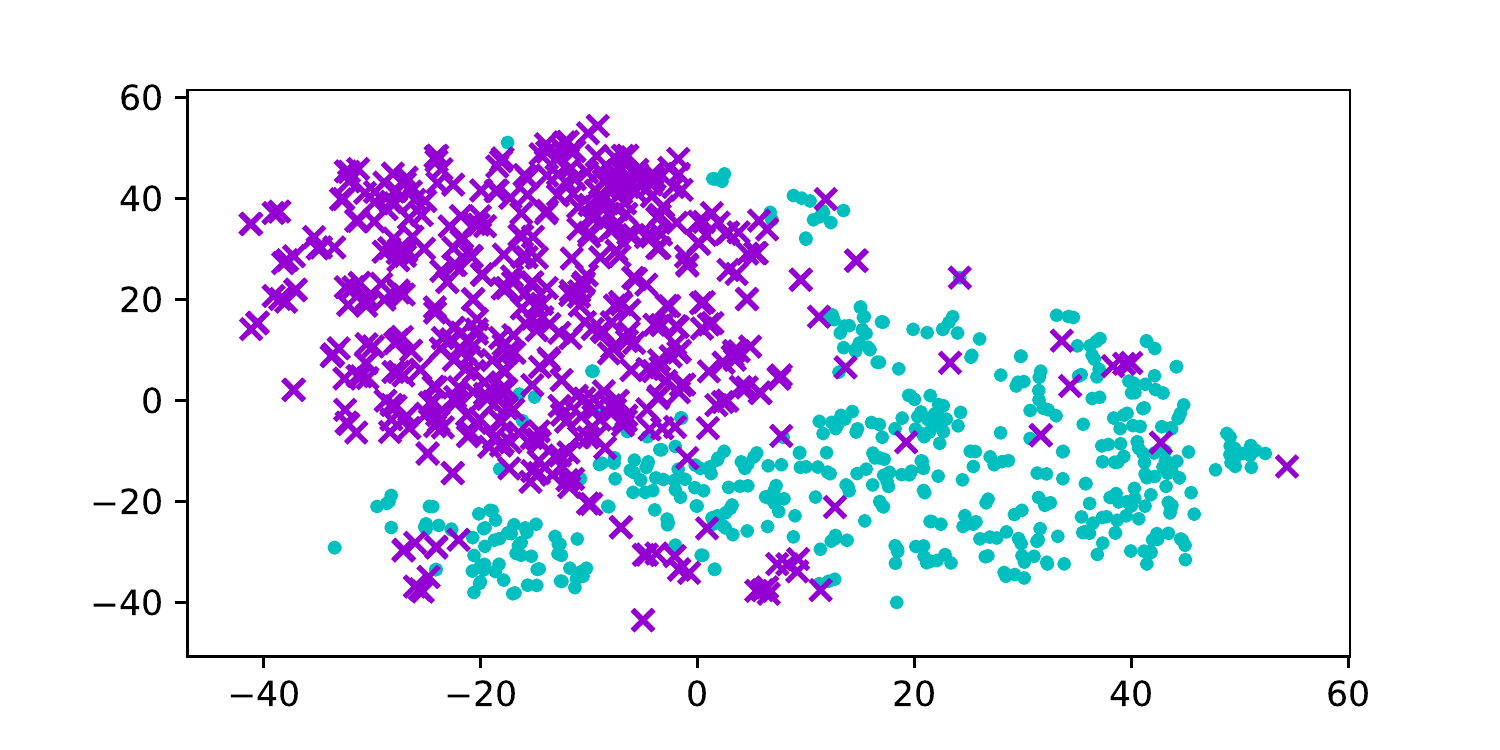}
  \caption{Hard-\wv}
  \label{fig:hard_wv}
\end{subfigure}
\begin{subfigure}{.46\textwidth}
  \centering
  \includegraphics[width=\linewidth]{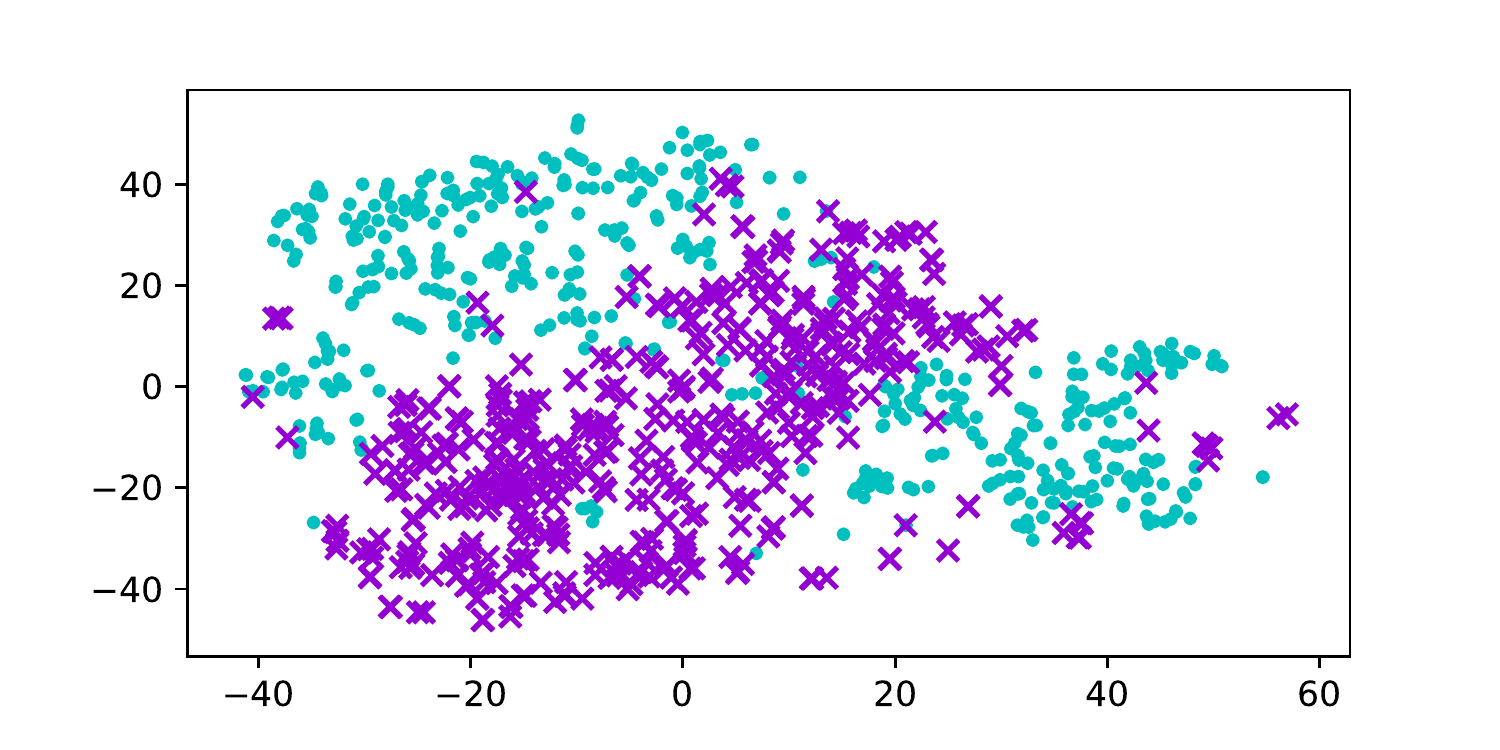}
  \caption{Double-Hard \wv}
  \label{fig:double_hd_wv}
\end{subfigure}%
\caption{tSNE visualization of top $500$ most male and female embeddings. Double-Hard \wv~mixes up two groups to the maximum extent, showing less gender information encoded.}
\label{fig:tsne_wv}
\end{figure}

\begin{table*}[!h]
\centering
\small
\begin{tabular}{ccccccc}
\toprule
\multirow{2}{*}{\textbf{Embeddings}} & \multicolumn{2}{c}{\bf Career \& Family } & \multicolumn{2}{c}{\bf Math \& Arts} & \multicolumn{2}{c}{\bf Science \& Arts}\\
 & \bf $d$ & \bf $p$ &  \bf $d$ & \bf $p$ & \bf $d$ & \bf $p$\\
\midrule
\wv & $1.89$ & $0.0$& $1.82$ & $0.0$ & $1.57$& $2\mathrm{e}^{-4}$\\
Hard-\wv & $1.80$ & $0.0$& $1.57$ & $7\mathrm{e}^{-5}$ & $0.83$& $0.05$\\
\midrule
Double-Hard \wv & $1.73$ & $0.0$& $1.51$ & $5\mathrm{e}^{-4}$ & $ 0.68$& $0.09$\\
\bottomrule
\end{tabular}
\caption{WEAT test of embeddings before/after Debiasing. The bias is insignificant when p-value, $p > 0.05$. Lower effective size ($d$) indicates less gender bias. Across all target words sets, Double-Hard \wv~ leads to the smallest effective size. Specifically, for Science \& Arts words, Double-Hard \wv~ successfully reaches a bias insignificant state ($p = 0.09$).}
\label{tab:WEAT_wv}
\end{table*}

\begin{table*}[!h]
\centering
\small
\begin{tabular}{c|cccc|cccc}
\toprule
\multirow{2}{*}{\bf Embeddings}& \multicolumn{4}{c|}{\bf Analogy} &\multicolumn{4}{c}{\bf Concept Categorization}
\\ 
 & Sem & Syn & Total & MSR & AP & ESSLI & Battig & BLESS\\
\midrule
\wv & $\bf 24.8$ & $\bf 66.5$ & $\bf55.3$ & $73.6$& $\bf 64.5$& $75.0$& $46.3$& $\bf 78.9$ \\
Hard-\wv  & $23.8$ & $66.3$ & $54.9$ & $73.5$& $62.7$& $75.0$& $\bf 47.1$& $77.4$\\
\midrule
Double-Hard \wv & $23.5$ & $66.3$ & $54.9$ & $\bf 74.0$& $63.2$& $75.0$& $46.5$& $77.9$\\
\midrule
\end{tabular}
\caption{Results of word embeddings on word analogy and concept categorization benchmark datasets. Performance (x100) is measured in accuracy and purity, respectively. On both tasks, there is no significant degradation of performance due to applying the proposed method.}
\label{tab:analogy_catrgorization_wv}
\end{table*}

We also apply \doublehd~on Word2Vec embeddings~\cite{word2vec} which have been widely used by many NLP applications. As shown in Figure~\ref{fig:discovering-w2v}, our algorithm is able to identify that the eighth principal component significantly affects the debiasing performance. 

Similarly, we first project away the identified direction $\mathbf{u}$ from the original \wv~embeddings and then apply Hard Debias algorithm. We compare embeddings debiased by our method with the original \wv~embeddings and Hard-\wv~ embeddings. 

Table~\ref{tab:bias_eval_wv} reports the experimental result using the neighborhood metric. Across three experiments where we cluster top $100$/$500$/$1000$ male and female words, Double-Hard \wv~ consistently achieves the lowest accuracy . Note that neighborhood metric reflects gender information that can be captured by the clustering algorithm. Experimental result validates that our method can further improve Hard Debias algorithm. This is also verified in Figure~\ref{fig:tsne_wv} where we conduct tSNE visualization of top $500$ male and female embeddings. While the original \wv~ embeddings clearly locate separately into two groups corresponding to different genders, this phenomenon  becomes less obvious after applying our debiasing method.

We further evaluate the debiasing outcome with WEAT test. Similar to experiments on \glove~ embeddings, we use male names and female names as attribute sets and analyze the association between attribute sets and three target sets. We report effective size and p-value in Table~\ref{tab:WEAT_wv}. Across three target sets, Double-Hard \wv~ is able to consistently reduce the effect size. More importantly, the bias related to Science \& Arts words becomes insignificant after applying our debiasing method.

To test the functionality of debiased embeddings, we again conduct experiments on word analogy and concept categorization tasks. Results are included in Table~\ref{tab:analogy_catrgorization_wv}. We demonstrate that our proposed debiasing method brings no significant performance degradation in these two tasks. 

To summarize, experiments on \wv~ embeddings also support our conclusion that the proposed Double-Hard Debiasing reduces gender bias to a larger degree while is able to maintain the semantic information in word embeddings.

%% file: acl2020.bbl
\begin{thebibliography}{27}
\expandafter\ifx\csname natexlab\endcsname\relax\def\natexlab#1{#1}\fi

\bibitem[{Almuhareb(2006)}]{DBLP:phd/ethos/Almuhareb06}
Abdulrahman Almuhareb. 2006.
\newblock \href {http://ethos.bl.uk/OrderDetails.do?uin=uk.bl.ethos.428974}
  {\emph{Attributes in lexical acquisition}}.
\newblock Ph.D. thesis, University of Essex, Colchester, {UK}.

\bibitem[{Baroni et~al.(2008)Baroni, Evert, and Lenci}]{essli}
Marco Baroni, Stefan Evert, and Alessandro Lenci. 2008.
\newblock Bridging the gap between semantic theory and computational
  simulations: Proceedings of the esslli workshop on distributional lexical
  semantics.

\bibitem[{Baroni and Lenci(2011)}]{Baroni:2011:WBD:2140490.2140491}
Marco Baroni and Alessandro Lenci. 2011.
\newblock \href {http://dl.acm.org/citation.cfm?id=2140490.2140491} {How we
  blessed distributional semantic evaluation}.
\newblock In \emph{Proceedings of the GEMS 2011 Workshop on GEometrical Models
  of Natural Language Semantics}, GEMS '11, pages 1--10, Stroudsburg, PA, USA.
  Association for Computational Linguistics.

\bibitem[{Basta et~al.(2019)Basta, Costa{-}juss{\`{a}}, and
  Casas}]{DBLP:journals/corr/abs-1904-08783}
Christine Basta, Marta~Ruiz Costa{-}juss{\`{a}}, and Noe Casas. 2019.
\newblock \href {http://arxiv.org/abs/1904.08783} {Evaluating the underlying
  gender bias in contextualized word embeddings}.
\newblock \emph{CoRR}, abs/1904.08783.

\bibitem[{Battig and Montague(1969)}]{Battig1969-BATCNO}
William~F. Battig and William~E. Montague. 1969.
\newblock \href {https://doi.org/10.1037/h0027577} {Category norms of verbal
  items in 56 categories a replication and extension of the connecticut
  category norms}.
\newblock \emph{Journal of Experimental Psychology}, 80(3p2):1.

\bibitem[{Bolukbasi et~al.(2016)Bolukbasi, Chang, Zou, Saligrama, and
  Kalai}]{Bolukbasi2016ManIT}
Tolga Bolukbasi, Kai-Wei Chang, James~Y. Zou, Venkatesh Saligrama, and
  Adam~Tauman Kalai. 2016.
\newblock Man is to computer programmer as woman is to homemaker? debiasing
  word embeddings.
\newblock In \emph{NIPS}.

\bibitem[{Caliskan et~al.(2017)Caliskan, Bryson, and Narayanan}]{Caliskan183}
Aylin Caliskan, Joanna~J. Bryson, and Arvind Narayanan. 2017.
\newblock \href {https://doi.org/10.1126/science.aal4230} {Semantics derived
  automatically from language corpora contain human-like biases}.
\newblock \emph{Science}, 356(6334):183--186.

\bibitem[{Chelba et~al.(2013)Chelba, Mikolov, Schuster, Ge, Brants, Koehn, and
  Robinson}]{chelba2013one}
Ciprian Chelba, Tomas Mikolov, Mike Schuster, Qi~Ge, Thorsten Brants, Phillipp
  Koehn, and Tony Robinson. 2013.
\newblock One billion word benchmark for measuring progress in statistical
  language modeling.
\newblock \emph{arXiv preprint arXiv:1312.3005}.

\bibitem[{Ethayarajh et~al.(2019)Ethayarajh, Duvenaud, and
  Hirst}]{ethayarajh2019understanding}
Kawin Ethayarajh, David Duvenaud, and Graeme Hirst. 2019.
\newblock Understanding undesirable word embedding associations.
\newblock \emph{arXiv preprint arXiv:1908.06361}.

\bibitem[{Gonen and Goldberg(2019)}]{Gonen2019LipstickOA}
Hila Gonen and Yoav Goldberg. 2019.
\newblock Lipstick on a pig: Debiasing methods cover up systematic gender
  biases in word embeddings but do not remove them.
\newblock In \emph{NAACL-HLT}.

\bibitem[{Gong et~al.(2018)Gong, He, Tan, Qin, Wang, and Liu}]{FRAGE}
Chengyue Gong, Di~He, Xu~Tan, Tao Qin, Liwei Wang, and Tie-Yan Liu. 2018.
\newblock \href
  {http://papers.nips.cc/paper/7408-frage-frequency-agnostic-word-representation.pdf}
  {Frage: Frequency-agnostic word representation}.
\newblock In S.~Bengio, H.~Wallach, H.~Larochelle, K.~Grauman, N.~Cesa-Bianchi,
  and R.~Garnett, editors, \emph{Advances in Neural Information Processing
  Systems 31}, pages 1334--1345. Curran Associates, Inc.

\bibitem[{Kaneko and Bollegala(2019)}]{gp_glove}
Masahiro Kaneko and Danushka Bollegala. 2019.
\newblock \href {http://arxiv.org/abs/1906.00742} {Gender-preserving debiasing
  for pre-trained word embeddings}.
\newblock \emph{CoRR}, abs/1906.00742.

\bibitem[{Kurita et~al.(2019)Kurita, Vyas, Pareek, Black, and
  Tsvetkov}]{BERTbias}
Keita Kurita, Nidhi Vyas, Ayush Pareek, Alan~W. Black, and Yulia Tsvetkov.
  2019.
\newblock \href {http://arxiv.org/abs/1906.07337} {Measuring bias in
  contextualized word representations}.
\newblock \emph{CoRR}, abs/1906.07337.

\bibitem[{Lee et~al.(2017)Lee, He, Lewis, and
  Zettlemoyer}]{DBLP:conf/emnlp/LeeHLZ17}
Kenton Lee, Luheng He, Mike Lewis, and Luke Zettlemoyer. 2017.
\newblock \href {https://www.aclweb.org/anthology/D17-1018/} {End-to-end neural
  coreference resolution}.
\newblock In \emph{Proceedings of the 2017 Conference on Empirical Methods in
  Natural Language Processing, {EMNLP} 2017, Copenhagen, Denmark, September
  9-11, 2017}, pages 188--197. Association for Computational Linguistics.

\bibitem[{van~der Maaten and Hinton(2008)}]{vanDerMaaten2008}
Laurens van~der Maaten and Geoffrey Hinton. 2008.
\newblock \href {http://www.jmlr.org/papers/v9/vandermaaten08a.html}
  {Visualizing data using {t-SNE}}.
\newblock \emph{Journal of Machine Learning Research}, 9:2579--2605.

\bibitem[{Manning et~al.(2008)Manning, Raghavan, and
  Schütze}]{manning2008introduction}
Christopher~D. Manning, Prabhakar Raghavan, and Hinrich Schütze. 2008.
\newblock \href
  {http://nlp.stanford.edu/IR-book/information-retrieval-book.html}
  {\emph{Introduction to Information Retrieval}}.
\newblock Cambridge University Press, Cambridge, UK.

\bibitem[{Mikolov et~al.(2013{\natexlab{a}})Mikolov, Chen, Corrado, and
  Dean}]{Mikolov2013EfficientEO}
Tomas Mikolov, Kai Chen, Gregory~S. Corrado, and Jeffrey Dean.
  2013{\natexlab{a}}.
\newblock Efficient estimation of word representations in vector space.
\newblock \emph{CoRR}, abs/1301.3781.

\bibitem[{Mikolov et~al.(2013{\natexlab{b}})Mikolov, Sutskever, Chen, Corrado,
  and Dean}]{word2vec}
Tomas Mikolov, Ilya Sutskever, Kai Chen, Greg~S Corrado, and Jeff Dean.
  2013{\natexlab{b}}.
\newblock Distributed representations of words and phrases and their
  compositionality.
\newblock In C.~J.~C. Burges, L.~Bottou, M.~Welling, Z.~Ghahramani, and K.~Q.
  Weinberger, editors, \emph{Advances in Neural Information Processing Systems
  26}, pages 3111--3119. Curran Associates, Inc.

\bibitem[{Mikolov et~al.(2013{\natexlab{c}})Mikolov, Yih, and
  Zweig}]{mikolov-etal-2013-linguistic}
Tomas Mikolov, Wen-tau Yih, and Geoffrey Zweig. 2013{\natexlab{c}}.
\newblock \href {https://www.aclweb.org/anthology/N13-1090} {Linguistic
  regularities in continuous space word representations}.
\newblock In \emph{Proceedings of the 2013 Conference of the North {A}merican
  Chapter of the Association for Computational Linguistics: Human Language
  Technologies}, pages 746--751, Atlanta, Georgia. Association for
  Computational Linguistics.

\bibitem[{Mu and Viswanath(2018)}]{mu2018allbutthetop}
Jiaqi Mu and Pramod Viswanath. 2018.
\newblock All-but-the-top: Simple and effective postprocessing for word
  representations.
\newblock In \emph{International Conference on Learning Representations}.

\bibitem[{Pennington et~al.(2014)Pennington, Socher, and
  Manning}]{pennington-etal-2014-glove}
Jeffrey Pennington, Richard Socher, and Christopher Manning. 2014.
\newblock {G}love: Global vectors for word representation.
\newblock In \emph{Proceedings of the 2014 Conference on Empirical Methods in
  Natural Language Processing ({EMNLP})}, pages 1532--1543, Doha, Qatar.
  Association for Computational Linguistics.

\bibitem[{Rudinger et~al.(2018)Rudinger, Naradowsky, Leonard, and
  Van~Durme}]{rudinger2018gender}
Rachel Rudinger, Jason Naradowsky, Brian Leonard, and Benjamin Van~Durme. 2018.
\newblock Gender bias in coreference resolution.
\newblock \emph{arXiv preprint arXiv:1804.09301}.

\bibitem[{Stanovsky et~al.(2019)Stanovsky, Smith, and
  Zettlemoyer}]{stanovsky2019evaluating}
Gabriel Stanovsky, Noah~A Smith, and Luke Zettlemoyer. 2019.
\newblock Evaluating gender bias in machine translation.
\newblock \emph{arXiv preprint arXiv:1906.00591}.

\bibitem[{Vig et~al.(2020)Vig, Gehrmann, Belinkov, Qian, Nevo, Singer, and
  Shieber}]{vig2020causal}
Jesse Vig, Sebastian Gehrmann, Yonatan Belinkov, Sharon Qian, Daniel Nevo,
  Yaron Singer, and Stuart Shieber. 2020.
\newblock \href {http://arxiv.org/abs/2004.12265} {Causal mediation analysis
  for interpreting neural nlp: The case of gender bias}.

\bibitem[{Zhao et~al.(2019)Zhao, Wang, Yatskar, Cotterell, Ordonez, and
  Chang}]{genderbiaselmo}
Jieyu Zhao, Tianlu Wang, Mark Yatskar, Ryan Cotterell, Vicente Ordonez, and
  Kai-Wei Chang. 2019.
\newblock Gender bias in contextualized word embeddings.
\newblock In \emph{North American Chapter of the Association for Computational
  Linguistics (NAACL)}.

\bibitem[{Zhao et~al.(2018{\natexlab{a}})Zhao, Wang, Yatskar, Ordonez, and
  Chang}]{ZWYOC18}
Jieyu Zhao, Tianlu Wang, Mark Yatskar, Vicente Ordonez, and Kai-Wei Chang.
  2018{\natexlab{a}}.
\newblock Gender bias in coreference resolution: Evaluation and debiasing
  methods.
\newblock In \emph{North American Chapter of the Association for Computational
  Linguistics (NAACL)}.

\bibitem[{Zhao et~al.(2018{\natexlab{b}})Zhao, Zhou, Li, Wang, and
  Chang}]{Zhao2018LearningGW}
Jieyu Zhao, Yichao Zhou, Zeyu Li, Wei Wang, and Kai-Wei Chang.
  2018{\natexlab{b}}.
\newblock Learning gender-neutral word embeddings.
\newblock In \emph{EMNLP}.

\end{thebibliography}
